\pdfoutput=1
\documentclass{article} 

\usepackage[preprint, nonatbib]{neurips_2024}

\usepackage{booktabs}
\usepackage{ulem}
\usepackage{graphicx} 
\usepackage[numbers]{natbib}
\usepackage{microtype}
\usepackage{hyperref}
\usepackage{url}
\usepackage{booktabs}
\usepackage{xspace}
\usepackage{lineno}
\usepackage{amsmath}
\usepackage{xcolor}
\usepackage{enumitem}

\usepackage{wrapfig}

\usepackage{caption}
\usepackage{adjustbox}

\usepackage[flushleft]{threeparttable}
\usepackage{enumitem}
\usepackage{verbatim}
\usepackage{listings}
\usepackage{amssymb}
\usepackage{scalerel} 
\usepackage{float}
\usepackage{subcaption}
\usepackage[inkscapeformat=png]{svg}
\usepackage{bbding}
\usepackage{comment}
\usepackage{tabularray}
\usepackage{booktabs,tabularx}
\usepackage{stfloats}

\newcommand{\method}{\texttt{PRIOR}\xspace}

\definecolor{ao(english)}{rgb}{0.0, 0.5, 0.0}

\NewDocumentCommand{\tong}
{ mO{} }{\textcolor{purple}{\textsuperscript{\textit{Tong}}\textsf{\textbf{\small[#1]}}}}

\NewDocumentCommand{\heng}
{ mO{} }{\textcolor{red}{\textsuperscript{\textit{Heng}}\textsf{\textbf{\small[#1]}}}}
\NewDocumentCommand{\hao}
{ mO{} }{\textcolor{purple}{\textsuperscript{\textit{Hao}}\textsf{\textbf{\small[#1]}}}}

\usepackage{soul}

\NewDocumentCommand{\yy}
{ mO{} }{\textcolor{pink}{\textsuperscript{\textit{coolYY}}\textsf{\textbf{\small[#1]}}}}

\NewDocumentCommand{\modi}
{ mO{} }{\textcolor{red}
{#1}}
\definecolor{navyblue}{RGB}{51,102,204}
\definecolor{highlightblue}{RGB}{135,206,250}
\sethlcolor{highlightblue}

\usepackage{microtype}
\hypersetup{
	colorlinks=true,
	linkcolor=cyan,
	filecolor=blue,      
	urlcolor=cyan,
	citecolor=purple,
}
\newcommand{\eref}[1]{Eq.~\ref{#1}}

\newcommand{\sref}[1]{\S\ref{#1}}
\newcommand{\fref}[1]{Fig.~\ref{#1}}

\title{Prioritizing Image-Related Tokens Enhances  \\ Vision-Language Pre-Training}

\author{Yangyi Chen, Hao Peng, Tong Zhang, Heng Ji\\ University of Illinois Urbana-Champaign \\  \texttt{yangyic3@illinois.edu} }

\begin{document}


\maketitle

\begin{figure*}[h!]
\centering
\vspace{-20pt}
\includegraphics[width=\textwidth]{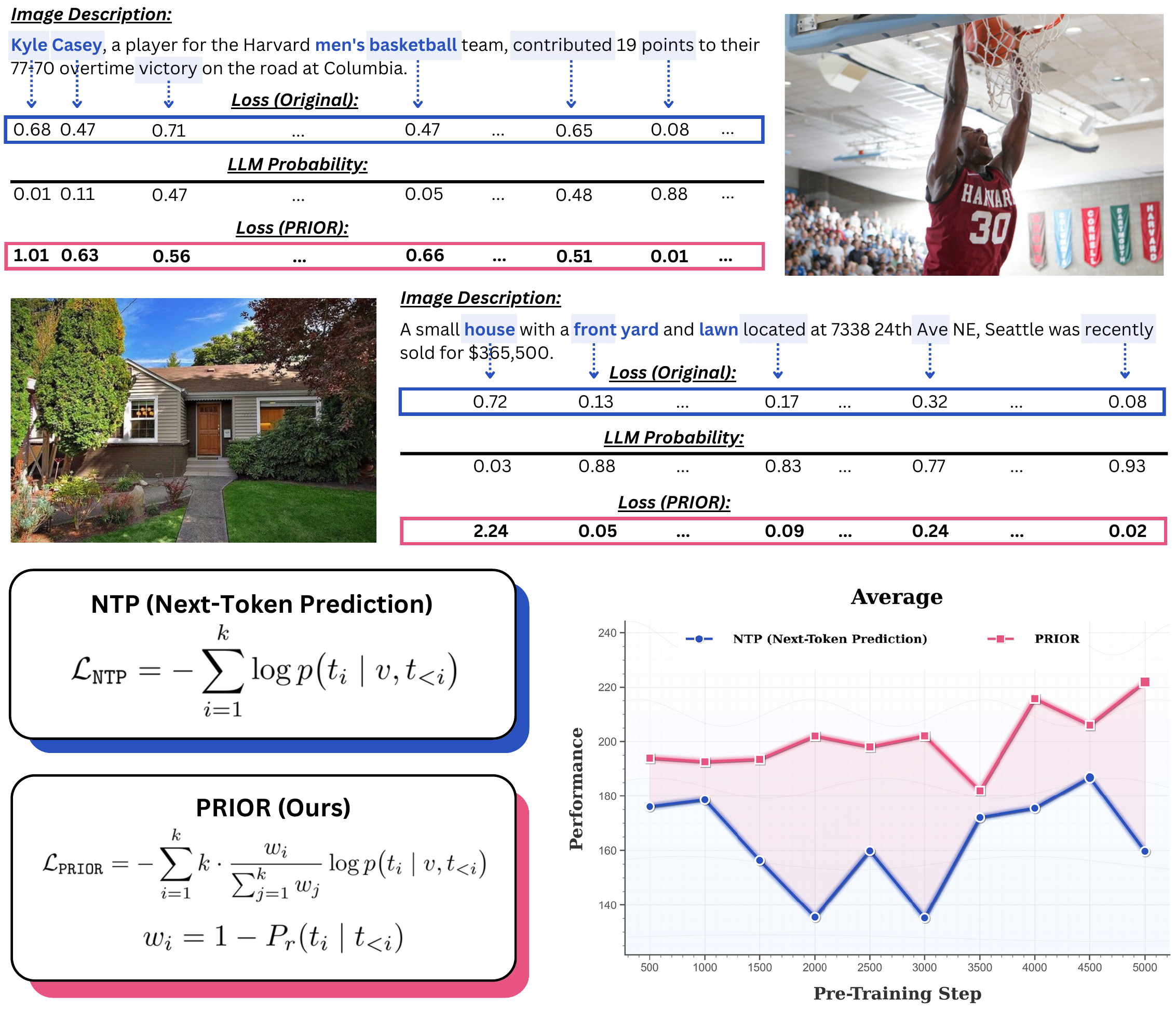}
 \caption{\textbf{(Top) Synthetic examples to highlight the motivation of \method.} 
 Only a few tokens in the captions (highlighted in \textcolor{navyblue}{blue}, word-level for better visualization) are related to the associated images. \method utilizes probability scores from a text-only LLM to recalibrate the original loss function at the token level, prioritizing image-related tokens that receive lower probability scores from the LLM.
 \textbf{(Bottom Left) \method formulation.} Given an image $v$ paired with a caption `$t_1, t_2, ..., t_k$', \method enhances vision-language pre-training by assigning a normalized weight to each token loss, which is computed based on the LLM probability $P_r(t_i | t_{<i})$. 
 \textbf{(Bottom Right) Performance of \method.} 
 \method demonstrates consistent performance improvement (average over several vision-language benchmarks) and better training stability compared to the widely used next-token prediction objective in vision-language pre-training.
 In addition, \method shows superior scaling behaviors in both performance predictability and potential improvement with increased compute and data (\sref{sec:sl}).
} 
 \label{fig:introfigure}
 \end{figure*}
 
\newpage

\begin{abstract}

In standard large vision-language models (LVLMs) pre-training, the model typically maximizes the joint probability of the caption conditioned on the image via next-token prediction (NTP); however, since only a small subset of caption tokens directly relates to the visual content, this naive NTP unintentionally fits the model to noise and increases the risk of hallucination. We present \textbf{\method}, a simple vision-language pre-training approach that addresses this issue by \textbf{prior}itizing image-related tokens through differential weighting in the NTP loss, drawing from the importance sampling framework. \method introduces a reference model—a text-only large language model (LLM) trained on the captions without image inputs, to weight each token based on its probability for LVLMs training. Intuitively, tokens that are directly related to the visual inputs are harder to predict without the image and thus receive lower probabilities from the text-only reference LLM. During training, we implement a token-specific re-weighting term based on the importance scores to adjust each token's loss. We implement \method in two distinct settings: LVLMs with visual encoders and LVLMs without visual encoders. We observe 19\% and 8\% average relative improvement, respectively, on several vision-language benchmarks compared to NTP. In addition, \method exhibits superior scaling properties, as demonstrated by significantly higher scaling coefficients, indicating greater potential for performance gains compared to NTP given increasing compute and data. The code will be available at \url{https://github.com/Yangyi-Chen/PRIOR}.

%
%

%


%
%

%

%
%

%

%

\end{abstract}

\section{Introduction}
%


Vision-language pre-training enhances both visual perception and visual-textual association capabilities in large vision-language models (LVLMs)~\citep{DBLP:conf/cvpr/ZhangLHY0WCG21, DBLP:journals/corr/abs-2412-10302}. 
However, our preliminary human annotations on 100 examples from Capsfusion~\cite{yu2024capsfusion} reveal that only 31.3\% of words in web-scale image-caption pairs directly relate to the associated images, with the remainder containing irrelevant information, stylistic elements, or website-specific content. 
For example, considering the second example in~\fref{fig:introfigure} (Top), only ``\textcolor{navyblue}{house}'', ``\textcolor{navyblue}{front yard}'', and ``\textcolor{navyblue}{lawn}'' are image-related, while the remaining tokens lack visual correspondence, such as the location and price information of the house.
The standard next-token prediction (NTP) objective treats all tokens equally, regardless of their relevance to visual content. 
Thus, besides learning visual perception and vision-language alignment, LVLMs inevitably model the entire caption distribution, potentially overfitting to noise and contributing to hallucination~\citep{DBLP:conf/acl/SoricutDSG18, DBLP:conf/cvpr/ChangpinyoSDS21, liu2023examining}.


\looseness=-1
In this work, we present \textbf{\method}, a simple approach that enhances vision-language pre-training by \textbf{prior}itizing optimization on image-related tokens.
\method addresses the above challenge by effectively distinguishing between image-related tokens and other tokens in the training corpus. 
We train a text-only reference model to capture the caption distribution without visual context in the vision-language corpus. 
This allows us to identify which tokens are likely to be image-related and establish an importance distribution over the token space. 
The key insight is that tokens easily predicted by the reference model likely contain minimal visual information, while those difficult for text-only models to predict more likely convey image-specific content.
We formalize this intuition by using the text-only model's predictive probability to reweight the NTP loss for each token, as shown in~\fref{fig:introfigure} (Bottom left). 
This training algorithm, inspired by importance sampling principles, prioritizes the optimization on tokens containing image-related information while preserving general language generation capabilities.

%

\method addresses several crucial limitations in previous vision-language pre-training approaches. Existing methods, as detailed in~\sref{sec:related}, primarily fall into two categories, each with their own challenges.
The first category involves supplementing NTP with additional training objectives such as contrastive loss~\citep{DBLP:journals/corr/abs-1908-03557} or distillation loss~\citep{liao2025multimodal},
but these approaches sacrifice NTP's simplicity and create challenges for large-scale, efficient implementation with existing frameworks~\citep{DBLP:conf/kdd/RasleyRRH20}.
In contrast, \method can be easily integrated with existing pre-training frameworks through minimal code changes for the loss computation and offline-computed token importance scores.
The second category focuses on optimizing training data through distillation~\citep{DBLP:conf/eccv/ChenLDZHWZL24},
filtering pipelines~\citep{DBLP:journals/corr/abs-2412-05237}, and related approaches.
While conceptually promising, these methods face inherent scalability constraints due to their dependence on teacher models or human-designed heuristics. 
In contrast, \method's text-only reference model offers a scalable alternative that dynamically extracts useful knowledge from noisy web-scale datasets, ensuring broad compatibility with advancements across vision-language datasets.

\looseness=-1
To evaluate \method's broad applicability across different LVLMs architectures, we conduct experiments with both vision encoder-based LVLMs~\citep{DBLP:journals/corr/abs-2304-10592, DBLP:journals/corr/abs-2301-12597} and end-to-end unified LVLMs with a simple linear projector for image processing~\citep{DBLP:journals/corr/abs-2407-06438, DBLP:conf/nips/DiaoCLWLW24, tao2024hovle}.
Our results demonstrate that \method consistently outperforms the naive NTP pre-training baseline, achieving 19\% and 8\% average relative improvement across these two architectures respectively, while also enhancing training stability. 
The average performance comparisons are visualized in~\fref{fig:introfigure} (Bottom Right) and~\fref{fig:avg_S}. 
Further analysis in \sref{sec:sl} reveals that \method exhibits superior scaling behaviors in terms of performance predictability (\fref{fig:sl}) and potential improvement with increased computational resources (\fref{fig:sl_2}). 
Moreover, as detailed in \sref{sec:loss_analysis},
\method accelerates pre-training by more effectively optimizing loss on both image-related and image-unrelated tokens.
%


%
%
%

\section{\method}

\subsection{Problem Formulation: Vision-Language Pre-training with Image-Caption Pairs}

We review the general vision-language pre-training that leverages a web-scale dataset comprising image-caption pairs ($v$, $c$) to develop models capable of generating relevant textual descriptions based on visual inputs. 
$v$ represents the image, and $c = t_1, t_2, ..., t_k$ represents the corresponding caption consisting of $k$ tokens.
The widely used vision-language pre-training objective, 
next-token prediction ($\mathcal{L_{\texttt{NTP}}}$), trains the model to predict each token in the caption based on the image and all previous tokens~\citep{DBLP:journals/corr/abs-2304-08485, DBLP:journals/corr/abs-2305-06500, DBLP:journals/corr/abs-2403-05525}, formally expressed as:
\begin{equation}
    \mathcal{L}_{\texttt{NTP}} = -\sum_{i=1}^{k} \log p\big(t_i \mid v, t_{<i}\big),
\label{eq:eq1}
\end{equation}
where $t_i$ is the current token to predict,
$t_{<i}$ represents the prefix.

%

%


\subsection{\method: Prioritizing Image-Related Tokens for Vision-Language Pre-Training}
\looseness=-1
\eref{eq:eq1} outlines a model that conditions token prediction on both images and text but lacks a mechanism to verify whether visual information is actually being effectively utilized.
%
The model can optimize this objective by relying exclusively on textual context, potentially resulting in LVLMs that overfit to supplementary text that doesn't correspond to visible image content.
To address this, we introduce \method, a simple method to advance the original vision-language pre-training by prioritizing image-related tokens, which are automatically identified by a text-only reference LLM.
Drawing from the importance sampling framework (\sref{sec:theory}), we use the reference model to construct a target distribution that assigns higher probability to tokens likely requiring visual information. This approach effectively concentrates optimization on image-related content while reducing emphasis on tokens predictable from text alone.


%
%

%

\paragraph{Text-Only LLM for Modeling Text Distribution}
\method introduces a text-only LLM as a reference to model the caption distribution without the image inputs. Specifically, we apply NTP loss exclusively on the text tokens, and the training objective is: 
\begin{equation}
    \mathcal{L}_{\texttt{TEXT}} = -\sum_{i=1}^{k} \log p_r\big(t_i \mid t_{<i}\big).
\label{eq:text}
\end{equation}
%

%

\paragraph{Vision-Language Pre-Training with Reweighted Tokens Loss}
\method utilizes the reference model's token probability to calculate the \textbf{importance score} $w_i$ for each token:
\begin{equation}
    w_{i} =  \big(1 - p_r(t_i \mid t_{<i})\big)^\alpha,
\label{eq:eq3}
\end{equation} 
where $\alpha$ (empirically set to 1) modulates the impact of $w_i$ during pre-training.
Tokens with higher importance scores are those that the text-only LLM finds difficult to predict, suggesting they are more likely to be image-related.
%
We implement this scoring process offline—computing and storing the importance score for each token beforehand—which eliminates the need for reference model inference during vision-language pre-training.

\method applies these token-specific importance scores to reweight the NTP loss. The resulting training objective is:
\begin{equation}
    \mathcal{L}_{\texttt{PRIOR}} = -\sum_{i=1}^{k} k \cdot \frac{w_i}{\sum_{j=1}^k w_j}  \log p\big(t_i \mid v, t_{<i}\big)
\label{eq:eq2}
\end{equation}
The normalization term $\tiny w_i/\sum_{j=1}^k w_j \normalsize$ ensures that the importance scores form a proper distribution across all $k$ tokens, while the multiplication by $k$ preserves the overall scale of the loss. 
This algorithm strategically upweights tokens that the text-only model struggles to predict, which typically correspond to image-specific information, while maintaining sufficient weight distribution across contextual tokens to ensure coherent language generation.




%

\subsection{\method as Importance Sampling}
\label{sec:theory}
\looseness=-1
We present the theoretical framework based on importance sampling that underpins and motivates \method.
In general, \method can be interpreted as a form of importance sampling where we draw more samples from regions where the reference model is uncertain.
In classical importance sampling, we estimate an expectation under a target distribution $p(x)$ using samples from a proposal distribution $q(x)$:

\begin{equation}
    \mathbb{E}_{p(x)}[f(x)] = \int f(x)p(x)dx = \int f(x)\frac{p(x)}{q(x)}q(x)dx = \mathbb{E}_{q(x)}\left[f(x)\frac{p(x)}{q(x)}\right],
\end{equation}
where the term $\frac{p(x)}{q(x)}$ represents the importance weight that corrects for the mismatch between the distributions.
For \method, we can formulate our problem as follows:
\begin{itemize}[noitemsep,topsep=0pt,parsep=3pt,partopsep=0pt,leftmargin=18pt]

\item \textbf{Function $f(x)$}:
The NTP loss (\textit{a.k.a,} negative log-likelihood loss) $-\log p_{\texttt{model}}(t_i | v, t_{<i})
$. 
    
\item \textbf{Sampling Distribution $q(x)$}: The empirical distribution in data $p_{\texttt{data}}(t_i | v, t_{<i})$.

\item \textbf{Target Distribution $p(x)$}: 
We aim to define a new target distribution $p_{\texttt{target}}(t_i | v, t_{<i})$ that assigns higher probability to tokens that are difficult for the reference model to predict:
\begin{equation}
    p_{\texttt{target}}(t_i | v, t_{<i}) \propto p_{\texttt{data}}(t_i | v, t_{<i}) \cdot \big(1-p_{\text{r}}(t_i | t_{<i})\big),
\label{eq:target}
\end{equation}
where $p_{\text{r}}(t_i | t_{<i})$ is the reference model's probability estimate. Intuitively, we upweight tokens that have low reference probability (surprising tokens). 
\end{itemize}

\textbf{Importance Sampling Formulation}:
Since we only have samples from $p_{\texttt{data}}(t_i | v, t_{<i})$, we use importance sampling to estimate the expectation under $p_{\texttt{target}}(t_i | v, t_{<i})$:
\begin{equation}
\begin{split}
    \mathbb{E}_{p_{\texttt{target}}(t_i | v, t_{<i})}[-\log p_{\texttt{model}}&(t_i | v, t_{<i})] = \\
    &\mathbb{E}_{p_{\texttt{data}}(t_i | i, t_{<i})}\left[-\log p_{\texttt{model}}(t_i | v, t_{<i}) \cdot \frac{p_{\texttt{target}}(t_i | v, t_{<i})}{p_{\texttt{data}}(t_i | v, t_{<i})}\right]
\end{split}
\label{eq:is}
\end{equation}

Thus, the importance weight for each token loss (\textit{i.e.,} $-\log p_{\texttt{model}}(t_i | v, t_{<i})$ in \eref{eq:is}), with the target distribution defined in~\eref{eq:target}, is:
\begin{equation}
w(t_i | v, t_{<i}) = \frac{p_{\texttt{target}}(t_i | v, t_{<i})}{p_{\texttt{data}}(t_i | v, t_{<i})} \propto \big(1-p_{\text{r}}(t_i | t_{<i})\big)
\end{equation}


To make this a proper probability distribution over the sequence, we normalize:
\begin{equation}
\tilde{w}(t_i | v, t_{<i}) = \frac{w(t_i | v, t_{<i})}{\sum_{j=1}^{k} {w}(t_j | v, t_{<j})}
\end{equation}
While this self-normalization introduces bias into our estimation of the expectation over the target distribution, it substantially reduces variance, yielding improved training stability—a well-established tradeoff 
in previous work~\citep{metelli2018policy, korbak2022rl}.
Our importance-sampled loss thus becomes:
\begin{equation}
L_{\text{IS}} = -\sum_{i=1}^{k} \tilde{w}_t(t_i | v, t_{<i}) \cdot \log p_{\texttt{model}}(t_i | v, t_{<i})    
\end{equation}
\looseness=-1
This weighted average represents the expected loss under our target distribution that emphasizes difficult tokens.
Note that this formulation is consistent with \eref{eq:eq2},
with the difference in the multiplication by $k$ to empirically keep the loss scale and stabilize the training. 
In addition, this importance sampling framework can be easily extended to consider the $\alpha$ term.
We also provide a theoretical justification for the key intuition of \method through mutual information analysis, as detailed in~\sref{app:mutual}. 

\section{Experiments}

\begin{figure}[t!]
  \centering
  \begin{minipage}[t]{0.33\textwidth}
    \centering
    \includegraphics[width=\textwidth]{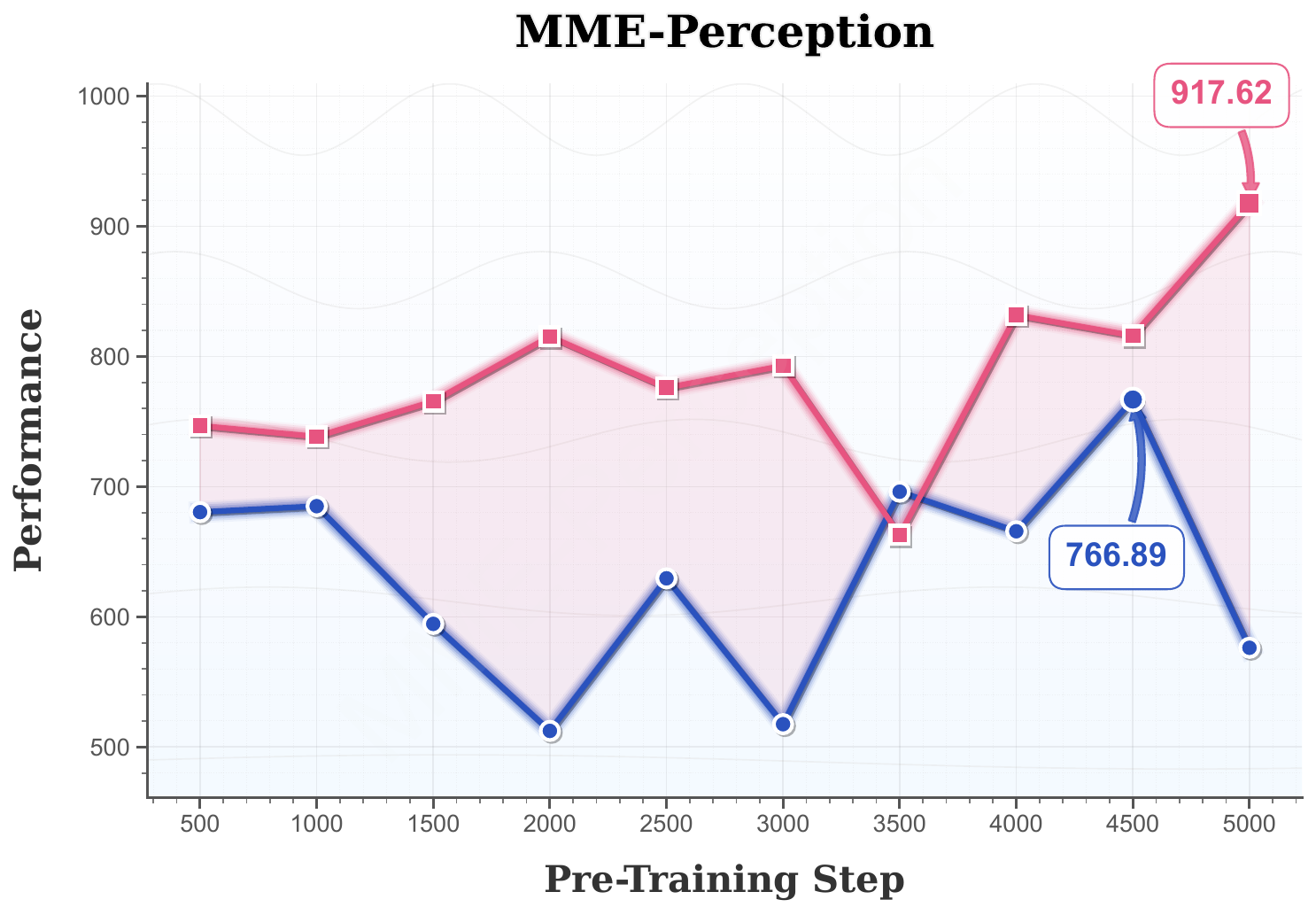}
    \label{fig:x1}
  \end{minipage}\hfill
  \begin{minipage}[t]{0.33\textwidth}
    \centering
    \includegraphics[width=\textwidth]{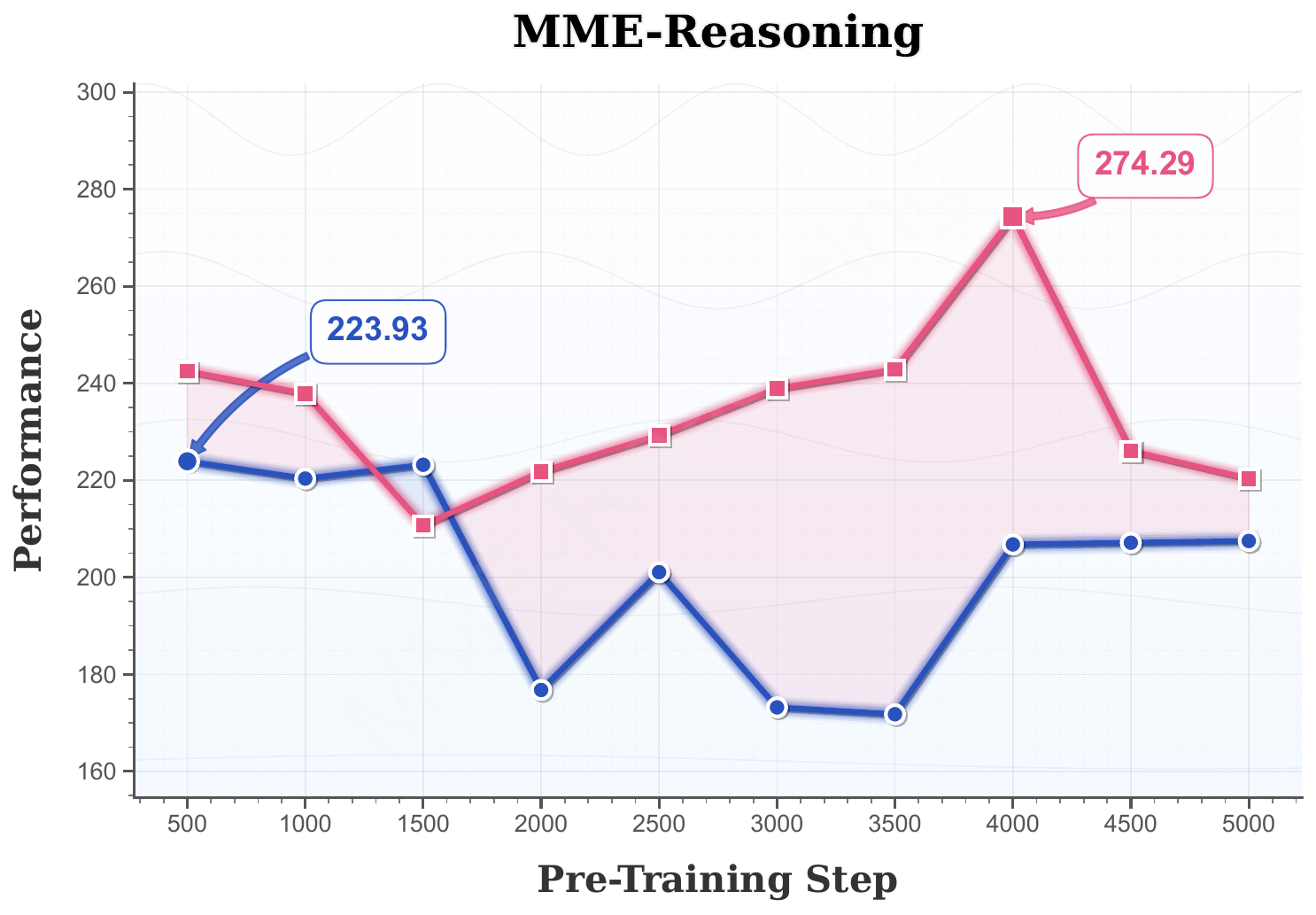}
    \label{fig:x2}
  \end{minipage}\hfill
  \begin{minipage}[t]{0.33\textwidth}
    \centering
    \includegraphics[width=\textwidth]{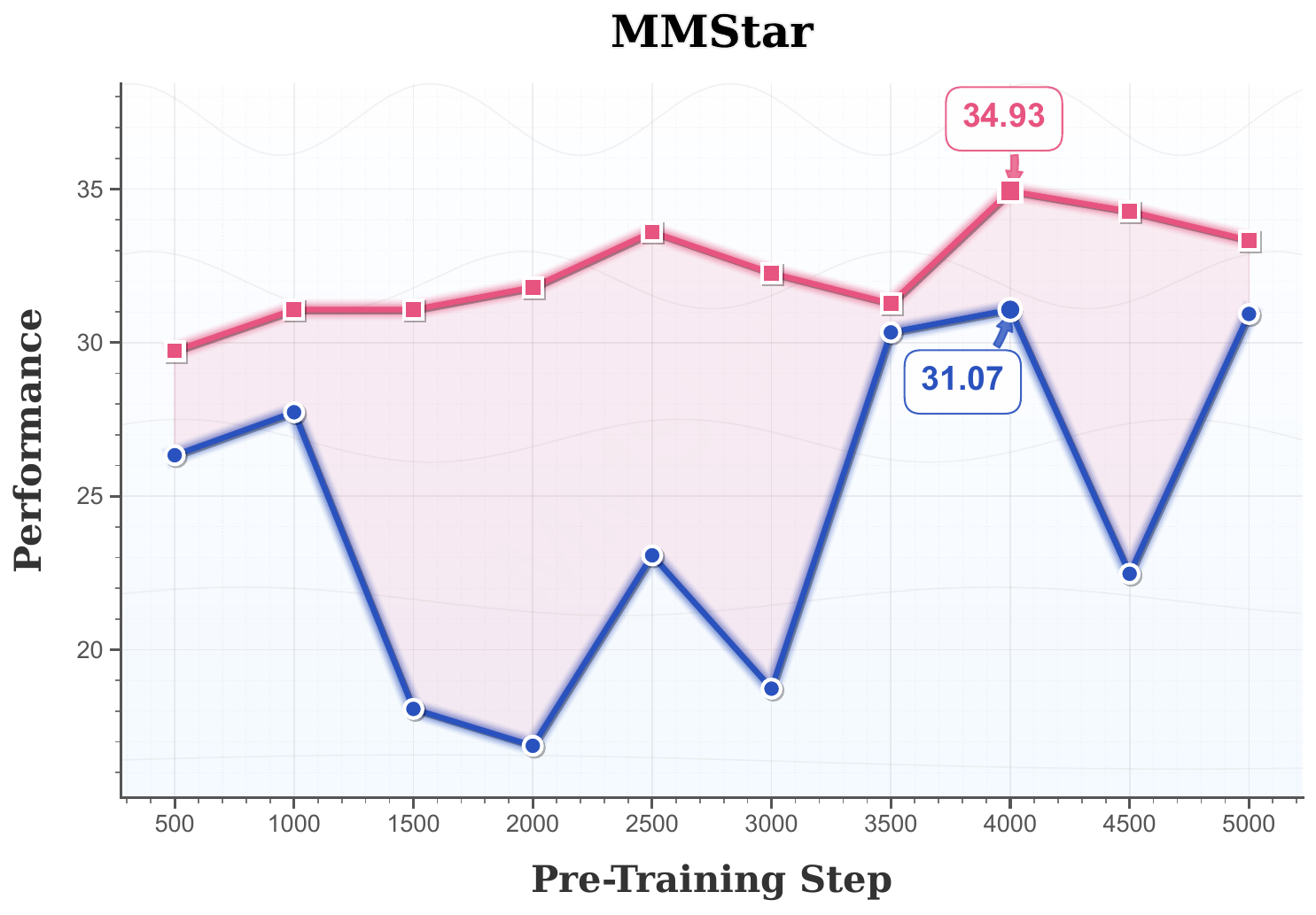}
    \label{fig:x3}
  \end{minipage}\hfill
  \vspace{-3pt}
  \begin{minipage}[t]{0.33\textwidth}
    \centering
    \includegraphics[width=\textwidth]{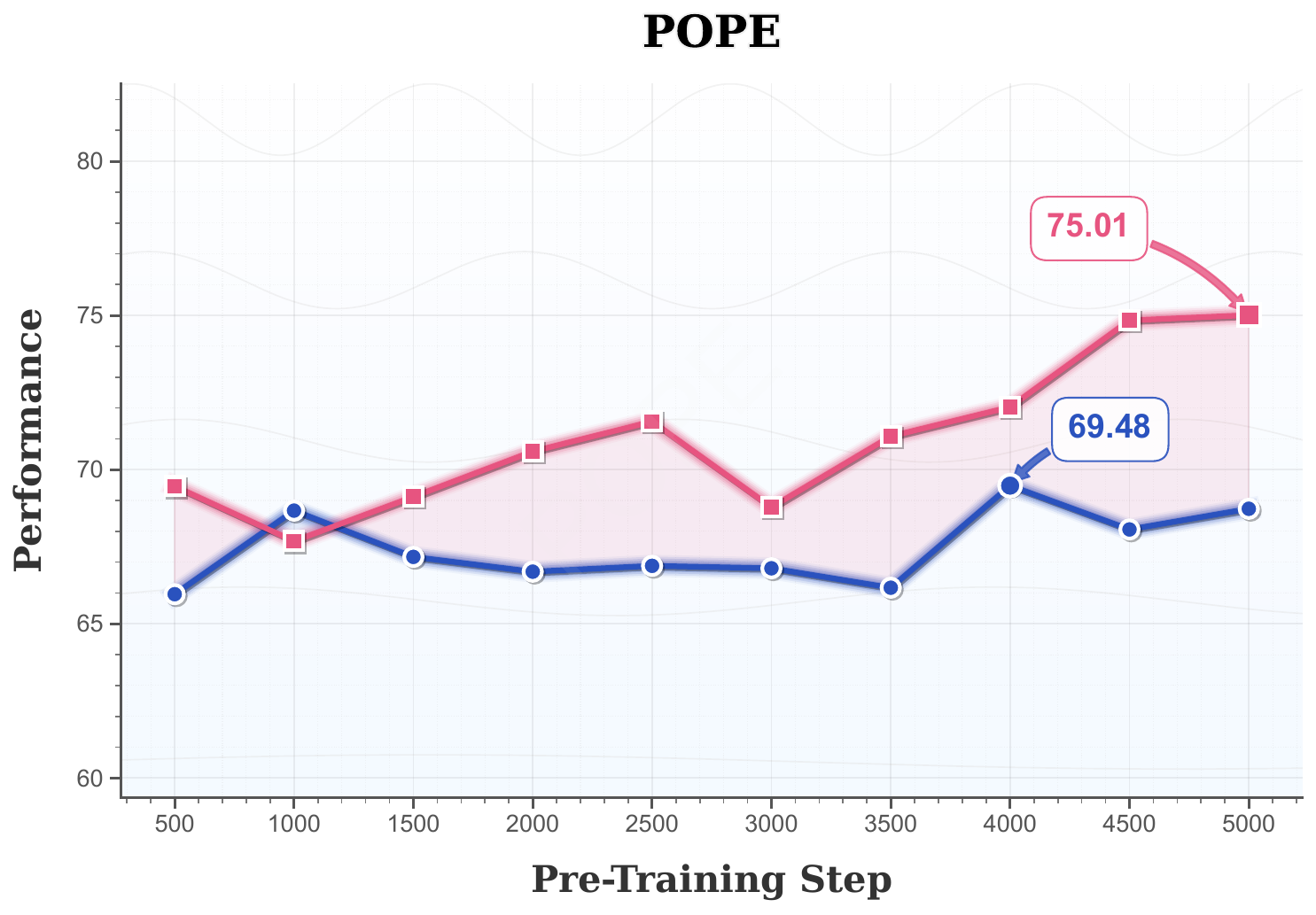}
    \label{fig:x1}
  \end{minipage}\hfill
  \begin{minipage}[t]{0.33\textwidth}
    \centering
    \includegraphics[width=\textwidth]{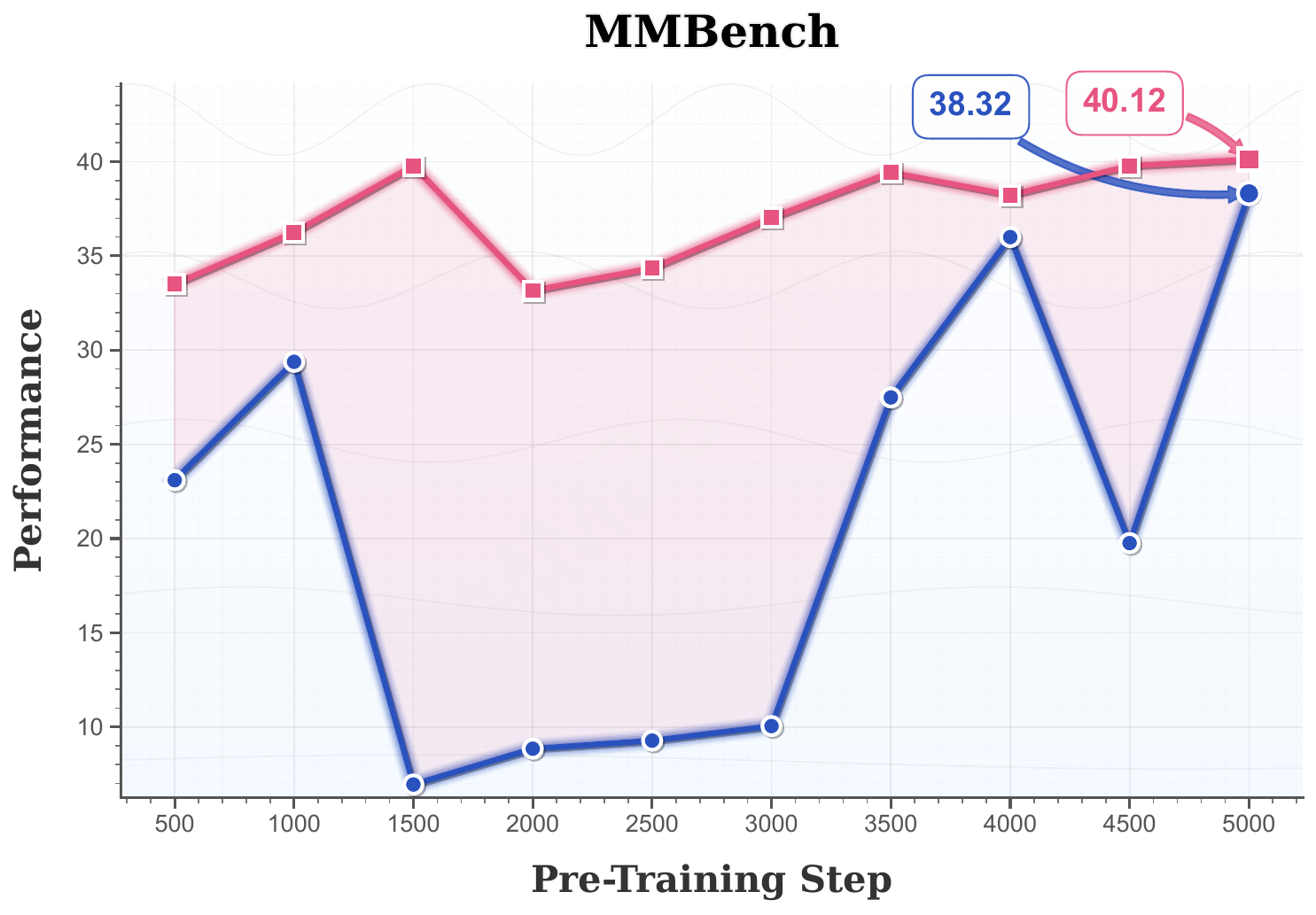}
  \end{minipage}\hfill
  \begin{minipage}[t]{0.33\textwidth}
    \centering
    \includegraphics[width=\textwidth]{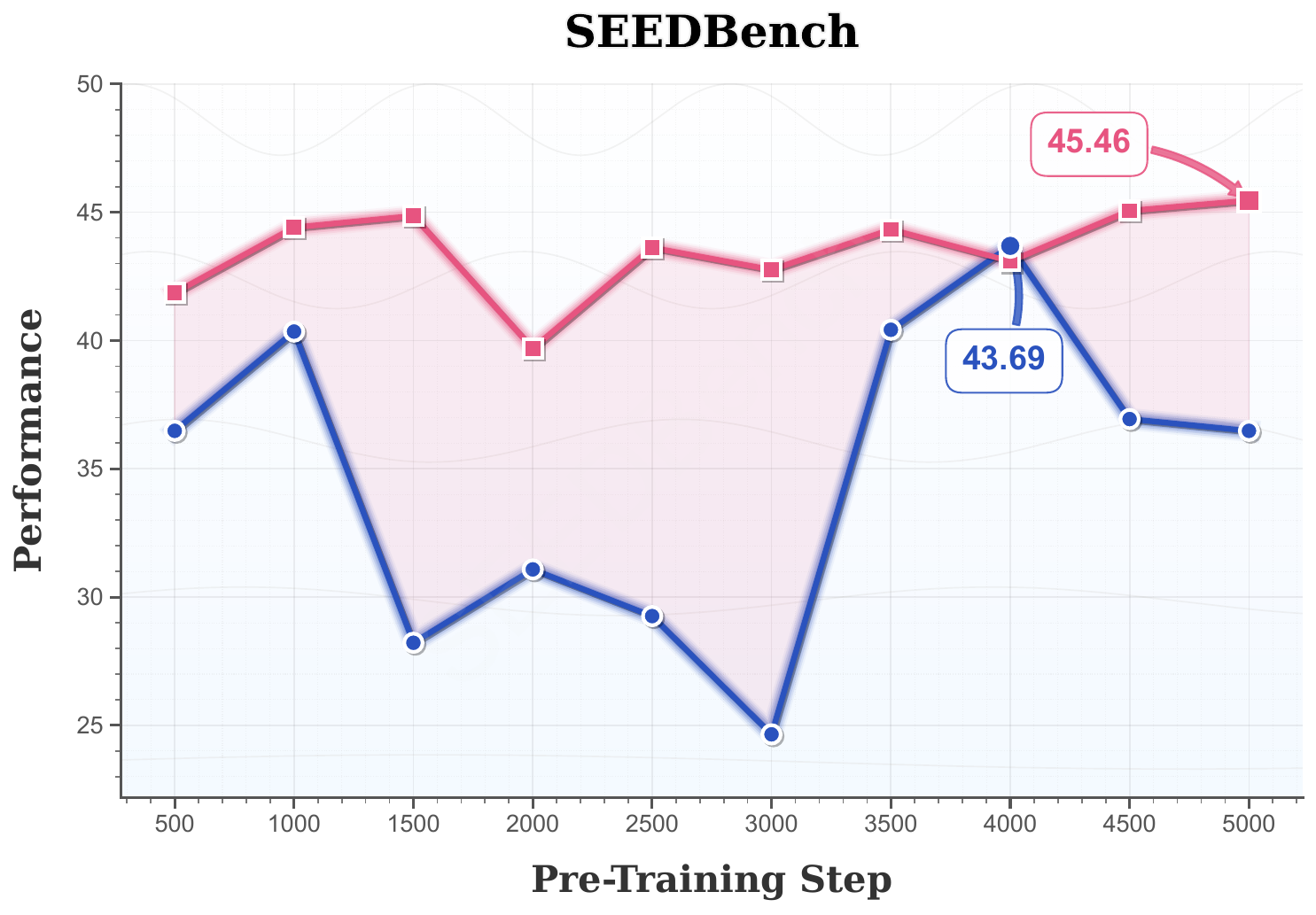}
  \end{minipage}\hfill
  \vspace{-10pt}
\includegraphics[width=0.45\textwidth]{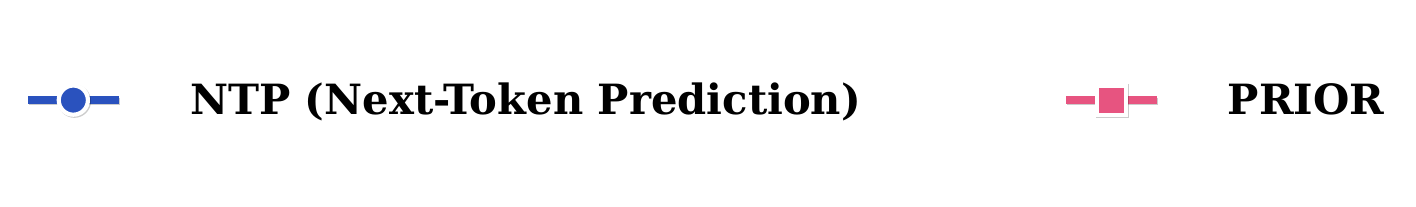}
  \vspace{-11pt}
\caption{\label{fig:main}\looseness=-1 \textbf{Main experimental results of LVLMs with pre-trained visual encoders.} 
We compare \method with the NTP vision-language pre-training across various training steps on LVLMs with pre-trained visual encoders, and we annotate the highest performance for each method, respectively. \method demonstrates both superior performance and greater stability throughout the entire training.}
\vspace{-7pt}
\end{figure}


\subsection{Implementation Details of \method}

We adopt CapsFusion ~\citep{yu2024capsfusion}, which comprises 120M image-text pairs for vision-language pre-training experiments.
We first sample a subset ($\tiny\sim$5M) and use only the caption to pre-train a text-only reference LLM (initialized with Llama-3-8B~\citep{DBLP:journals/corr/abs-2407-21783}). 
Since the captions in vision-language datasets are typically shorter, we pack multiple samples within the context length (8192) for efficient pre-training.
%

For vision-language pre-training, we sample another subset ($\tiny\sim$3M) and compute token-level reference probabilities $p_r(t_i|t_{<i})$ offline using the reference LLM. We store this data as (image, caption, token-level reference probability list) tuples.
To measure the extensive applicability of \method across diverse LVLMs architectures, we investigate two types of LVLMs:
\begin{itemize}[noitemsep,topsep=0pt,parsep=3pt,partopsep=0pt,leftmargin=18pt]
\item \textbf{H-LVLMs}: We implement LVLMs with pre-trained visual encoders (\textit{i.e.,}\textbf{H}eterogeneous architectures). 
Following the typical LLaVA-Style LVLMs design~\citep{DBLP:journals/corr/abs-2304-08485}, we adopt a straightforward ViT-MLP-LLM architecture, employing Llama-3.2-3B as the LLM backbone for efficient experimentation and CLIP-ViT-Large-336~\citep{radford2021learning} as the vision encoder.
Following~\citet{liu2023improved}, we only train the MLP component during pre-training to learn the vision-language alignment and maintain consistent language capabilities across all LVLMs for our controlled study.

\item \textbf{U-LVLMs}: We implement LVLMs with \textbf{U}nified architectures, which drop the pre-trained visual encoders and adopt a single end-to-end Transformer architecture for vision-language modeling~\citep{DBLP:journals/corr/abs-2407-06438, lei2025scalability, zhang2025pixel}.
Following~\citet{DBLP:journals/corr/abs-2407-06438}, we initialize with Llama-8B and pre-train it on ImageNet for 500 steps before further pre-training on image-caption data.

\end{itemize}
For the two types of LVLMs, we set the total training steps as 5,000, record every 500 steps, with the batch size as 512. 
During U-LVLM pre-training, we also incorporate an equal proportion of DCLM language dataset~\citep{li2024datacomp} to maintain general language capabilities.

\subsection{Experimental Setting}


\paragraph{Evaluation Benchmarks}
We select the following benchmarks for evaluation:
(1) MME~\citep{fu2024mme}, which measures both the perception and reasoning capabilities. Thus, we report MME-Perception and MME-Reasoning respectively.
(2) MMStar~\citep{chen2024we}, which measures advanced vision-language skills. 
(3) POPE~\citep{li2023evaluating}, a benchmark for object hallucination evaluation. 
(4) MMBench~\citep{liu2024mmbench}, a benchmark designed for general vision-language capacities evaluation. 
(5) SEEDBench~\citep{li2024seed}, which covers 12 evaluation dimensions including diverse aspects of LVLMs. 


\paragraph{Evaluation Setting} 
To elicit meaningful responses beyond image captions during evaluation, we conduct 20 steps for instruction fine-tuning on all pre-trained checkpoints using the same data subset from the post-training set in~\citet{DBLP:journals/corr/abs-2407-06438}.
We employ VLMEvalKit~\citep{ duan2024vlmevalkit} for unified evaluation and comparison across models.


\begin{figure}[t!]
  \centering
 \begin{minipage}[b]{0.48\textwidth}
    \centering
    \includegraphics[width=\textwidth]{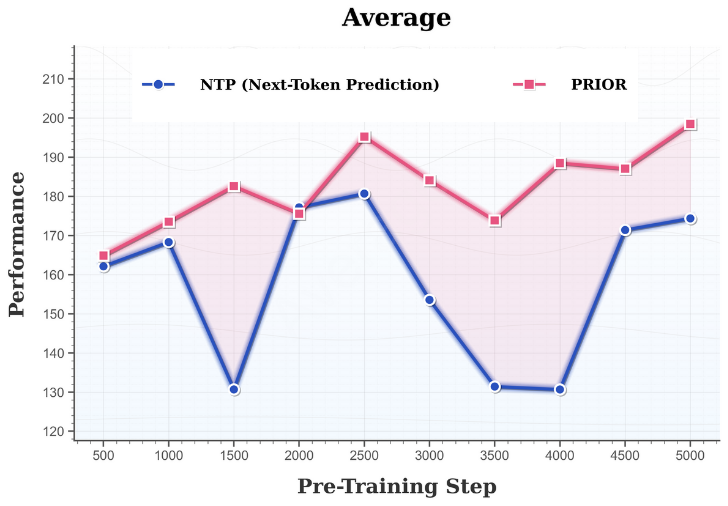}
    \caption{\label{fig:avg_S} \textbf{The average performance comparison on LVLMs with unified architectures.} \method demonstrates better performance and stability across the entire training process.}
      \vspace{-16pt}
  \end{minipage}
  \hfill
   \begin{minipage}[b]{0.48\textwidth}
    \centering
    \includegraphics[width=\textwidth]{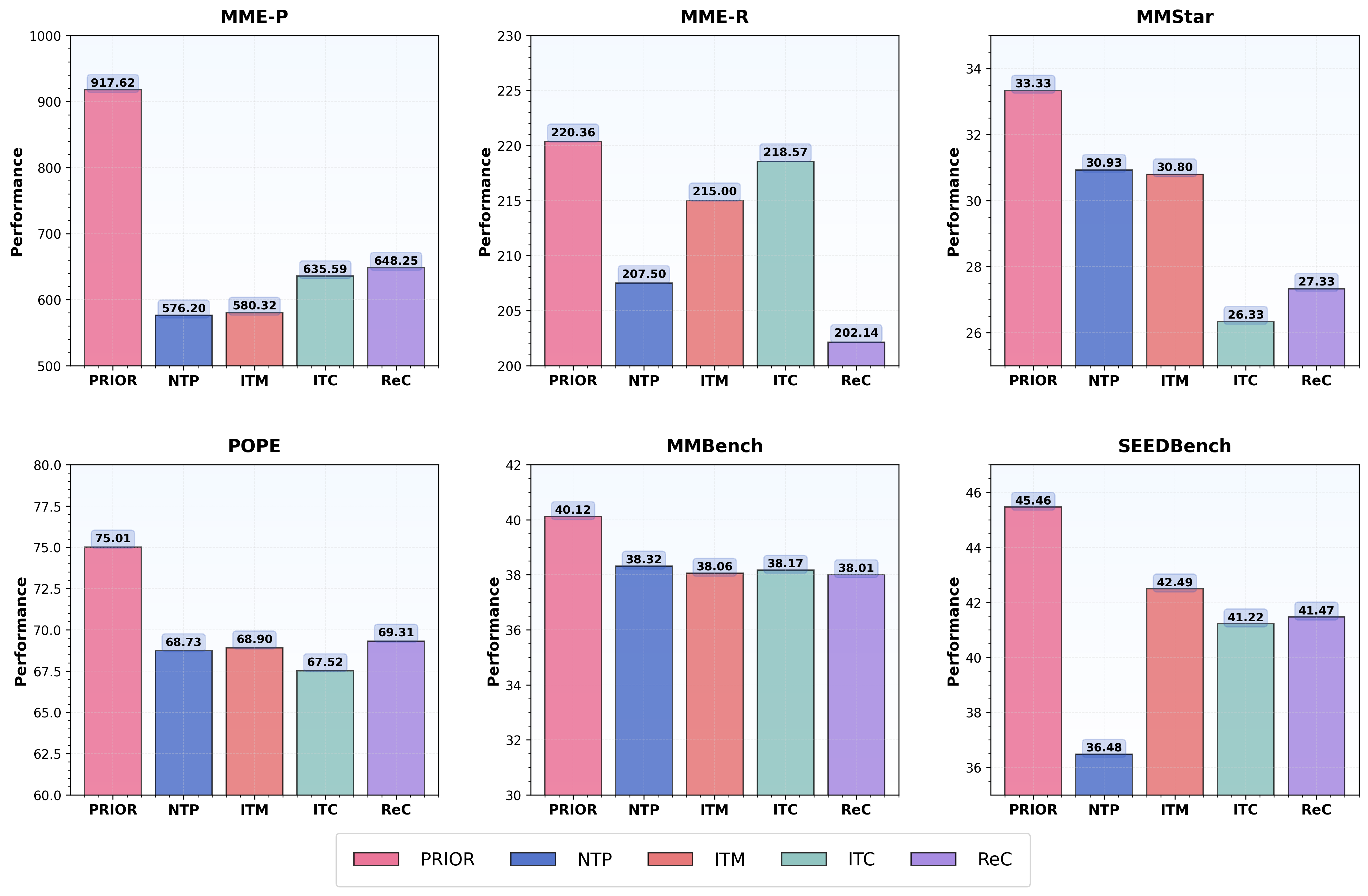}
    \caption{\label{fig:baseline}\textbf{The comparison with additional baselines on H-LVLMs with 5,000 training steps.}  \method consistently performs better and is simple to implement.}
    \vspace{-16pt}
  \end{minipage}
\end{figure}
\subsection{Experimental Results}
The main experimental results for each dataset are shown in~\fref{fig:main} (H-LVLMs) and~\fref{fig:main_S},~\sref{app:additional_exps} (U-LVLMs).
 Average performance across all datasets is summarized in~\fref{fig:introfigure}, bottom right (H-LVLMs) and~\fref{fig:avg_S} (U-LVLMs).
We have the following findings: 
\begin{itemize}[noitemsep,topsep=0pt,parsep=5pt,partopsep=0pt,leftmargin=18pt]
\item \textbf{\method generally outperforms the naive NTP pre-training objective by a large margin.}
We observe that \method enhances performance across the entire training trajectory for both H-LVLMs and U-LVLMs, demonstrating improvements compared to the NTP objective at most intermediate training steps.
For fully converged models (\textit{i.e.,} at 5,000 steps), we observe average relative improvements across all datasets of 18.61\% for H-LVLMs and 7.93\% for U-LVLMs.

\item \textbf{\method demonstrates better training stability.} 
We observe performance fluctuations during training across both H-LVLMs and U-LVLMs. The NTP objective causes significant performance drops in MMBench and SEEDBench for H-LVLMs, while producing sharp performance fluctuations in POPE and MMBench for U-LVLMs. 
\method significantly reduces these fluctuations, demonstrating superior training stability compared to the naive NTP pre-training objective.

\item \textbf{\method exhibits higher potential in performance.} 
The highest performance scores are highlighted in~\fref{fig:main} and~\fref{fig:main_S} for each dataset. 
 Our results show \method outperforms the NTP objective in peak performance for both LVLM types. For H-LVLMs specifically, \method achieves optimal performance in final checkpoints, while NTP peaks earlier in training. This suggests \method benefits from extended training, likely due to its comprehensive optimization approach that continues refining multimodal representations in later stages. \method's sustained improvement trajectory indicates greater performance potential.


\end{itemize}

\subsection{Comparison with Additional Baselines}
We compare \method with the following approaches specifically designed for H-LVLMs:
(1) \textbf{ITM}: Image-text matching objective that pre-trains the MLP connector to map the $[\texttt{CLS}]$ token embedding close to the caption embedding \citep{DBLP:journals/corr/abs-2301-12597, chen2024internvl}.
(2) \textbf{ITC}: Image-text contrastive learning objective that pre-trains the MLP connector to align projected image representations with text representations \citep{chen2024internvl, radford2021learning}.
(3) \textbf{ReC}: Reconstructive tuning objective that supervises LVLMs to reconstruct images, focusing on the inherent richness and detail within input images \citep{DBLP:journals/corr/abs-2410-09575}.
All these methods are combined with NTP for 5,000 training steps.

The results presented in \fref{fig:baseline} demonstrate that \method consistently outperforms these baselines across all benchmarks, with particularly significant gains observed on complex reasoning tasks (\textit{i.e.,} MME-R, MMStar).
Additionally, \method integrates more seamlessly with existing pre-training frameworks, requiring only modifications to the loss computation when data is stored offline. 
This enables scalable pre-training of LVLMs on large-scale datasets across compute clusters.

\begin{figure}[t!]
  \centering
  \includegraphics[width=\textwidth]{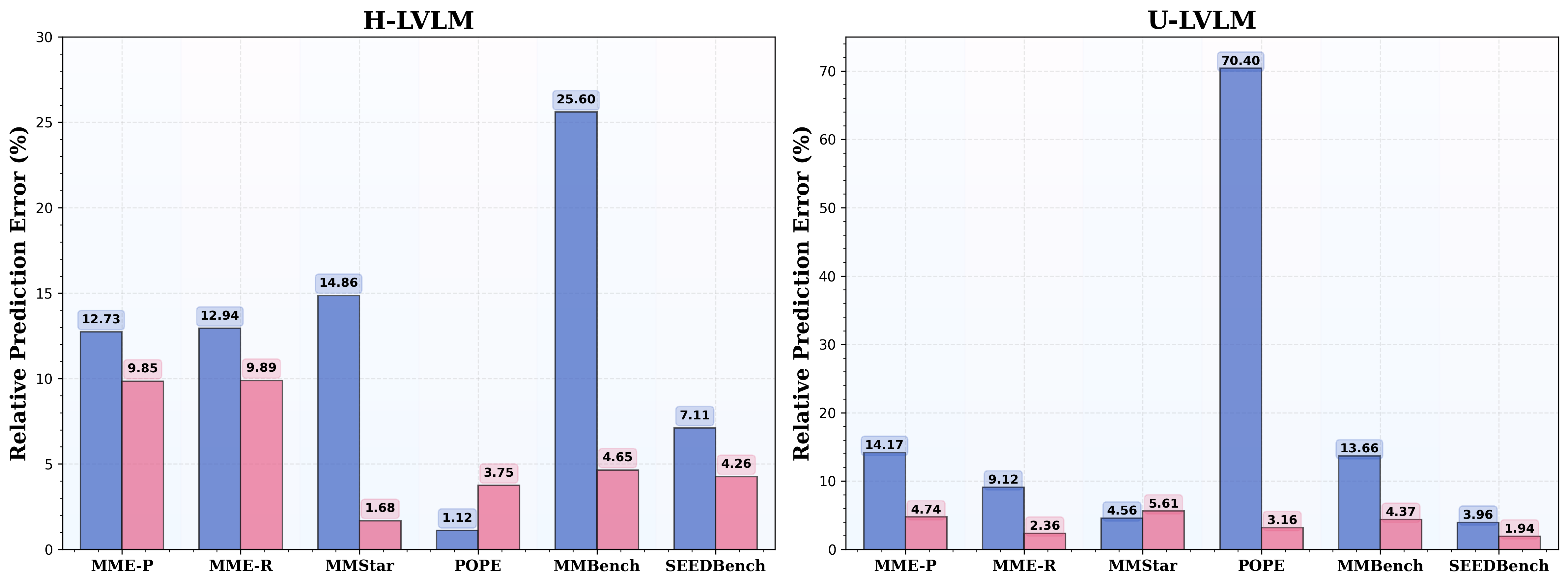}
\includegraphics[width=0.45\textwidth]{figs/raw_figs/elegant_legend.pdf}
  \vspace{-8pt}
\caption{\label{fig:sl}\textbf{The relative prediction error of NTP and \method.} We observe that the performance of LVLMs trained via \method is more predictable at scale.}
\vspace{-8pt}
\end{figure}

\begin{figure}[t!]
  \centering
  \includegraphics[width=\textwidth]{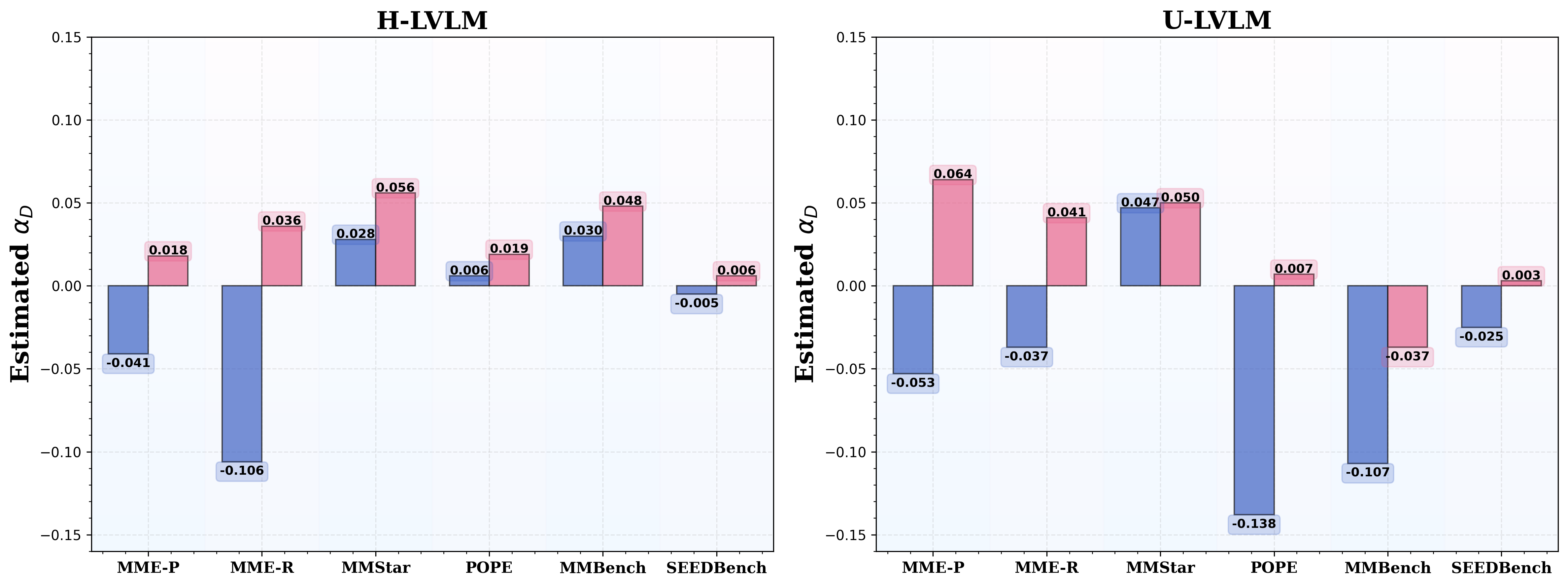}
\includegraphics[width=0.45\textwidth]{figs/raw_figs/elegant_legend.pdf}
  \vspace{-8pt}
\caption{\label{fig:sl_2}\textbf{The scaling behavior comparison of NTP and \method.} \method shows better scaling coefficients, indicating higher efficiency in translating increased resources into performance gains.}
\vspace{-10pt}
\end{figure}

\subsection{Scaling Laws Analysis: \method Scales Predictably and Reliably with Increasing Compute}
\label{sec:sl}
A fundamental distinction exists between LLMs and LVLMs regarding pre-training objectives. 
While NTP has proven to be an effective proxy for downstream task performance in LLMs~\citep{chen2024scaling, huang2024compression}, this correlation does not extend to LVLMs~\citep{DBLP:journals/corr/abs-2407-06438}.
Specifically, optimizing LVLMs to better predict caption given images through NTP does not reliably improve performance on downstream benchmarks.
In this section, we conduct a critical analysis of the scaling behaviors of NTP and \method. 
We train both H-LVLMs and U-LVLMs using different amounts of data (ranging from 7M to 70M training tokens, and including 8 sampling models), and use the following analytical form to estimate the scaling laws of NTP and \method~\citep{kaplan2020scaling}: 
\begin{equation}
L(D) = \left( \frac{D}{D_c} \right)^{\alpha_D},
\end{equation}
where $\alpha_D$ and $D_c$ are constants to be estimated, $D$ is the amounts of token used in training, and $L$ is the predictive performance.
This analytical form differs slightly from~\citet{kaplan2020scaling} since we are targeting the benchmark performance rather than the pre-training loss. A larger $\alpha_D$ indicates superior scaling behavior, reflecting greater performance improvement for an equivalent amount of training tokens. 
With the fitted analytical function, we investigate two problems:
\begin{itemize}[noitemsep,topsep=0pt,parsep=5pt,partopsep=0pt,leftmargin=18pt]
\item  \textbf{Predictability of model performance:} We use the fitted function to estimate the performance of LVLMs trained on 100M tokens, and measure the relative prediction error: 
\begin{equation}
\text{\small Relative Prediction Error} = \frac{| \text{\small Predictive Performance} - \text{\small Actual Performance} |}{\text{\small Actual Performance}}
\end{equation}
The results are presented in~\fref{fig:sl}.
The downstream performance of both H-LVLMs and U-LVLMs exhibits significantly higher predictability when trained using \method.
This property facilitates more reliable performance estimation for larger-scale deployments, addressing a critical challenge in production-level LVLM implementation

\item \textbf{Scaling behavior:} \fref{fig:sl_2} illustrates the scaling factor (\textit{i.e.,} $\alpha_D$) comparison for both methods. 
\method demonstrates better scaling properties across two LVLMs architectures, consistently achieving higher $\alpha_D$ than NTP. 
This indicates that \method more efficiently converts additional training data and compute into improved downstream performance.

\end{itemize}

\section{Further Analysis}


%

%


\begin{figure}[t!]
  \centering
  \begin{minipage}[b]{0.48\textwidth}
    \centering
    \includegraphics[width=\textwidth]{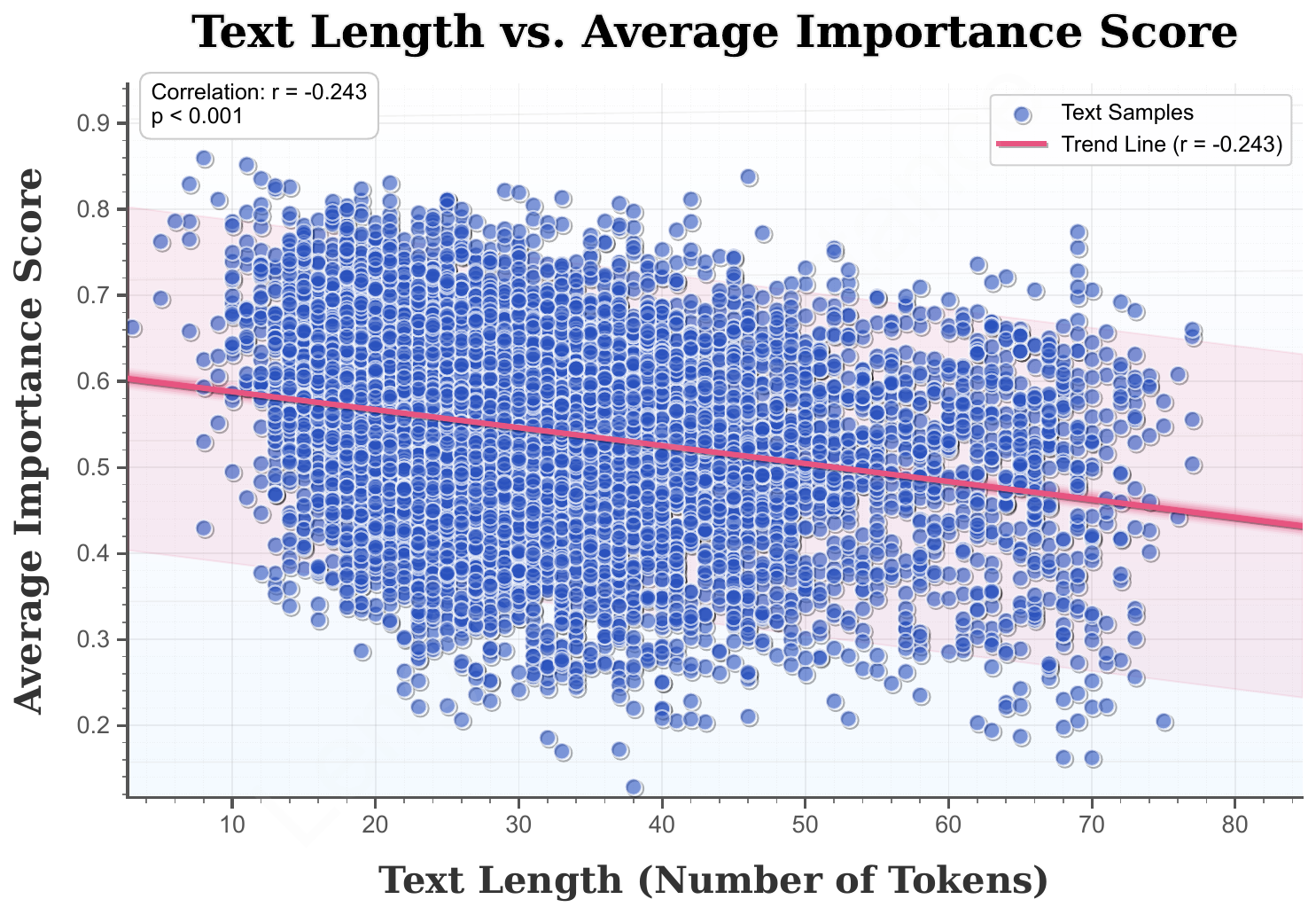}
   
  \end{minipage}
  \hfill
  \begin{minipage}[b]{0.48\textwidth}
    \centering
    \includegraphics[width=\textwidth]{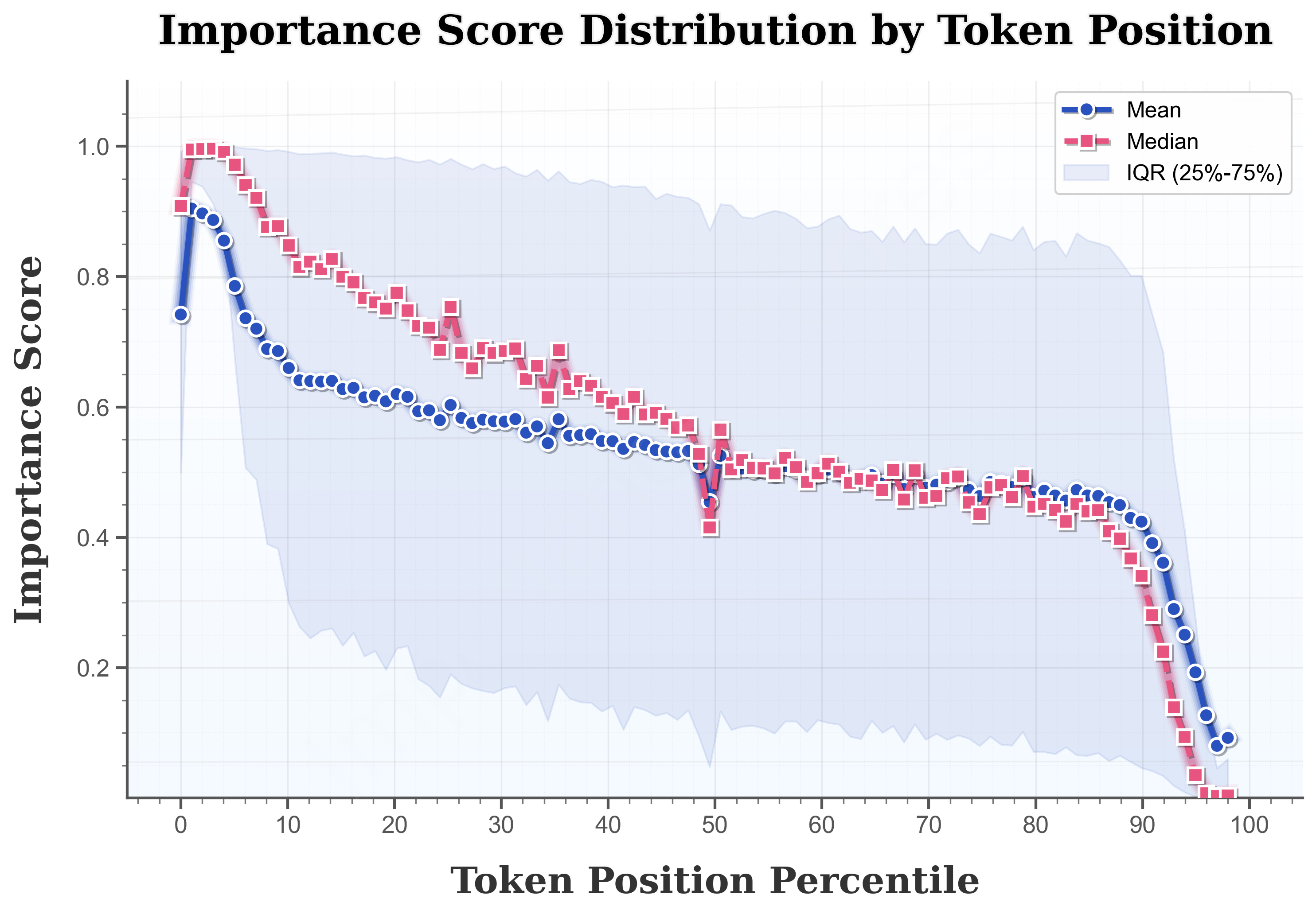}
  \end{minipage}
  \caption{\label{fig:quantitative_analysis}\textbf{Quantitative analysis of the importance score distribution.} We find that the average importance score for each caption decreases (linearly) with the text length. Within each caption, the importance score decreases in later positions.}
  \vspace{-18pt}
\end{figure}
\subsection{Importance Score Distribution}
\method adjusts the loss for each token based on the importance score $w_i$ assigned to that token. 
We conduct a quantitative analysis to understand the distribution of the importance score. 
The results are presented in~\fref{fig:quantitative_analysis}. 
We examine the relation between text length (\textit{i.e.,} number of tokens) and the average importance score across all tokens within the caption, finding that the average importance score decreases as text length increases, with a linear correlation of $r$=-0.243. 
This suggests longer caption introduces more information absent from images, potentially including background content. 
Analysis of importance scores by token position reveals that initial tokens receive high importance scores, with scores progressively declining for later tokens. This indicates that later tokens can be more readily predicted from previous tokens without image reference.


\begin{figure}[t!]
  \centering
  \includegraphics[width=\textwidth]{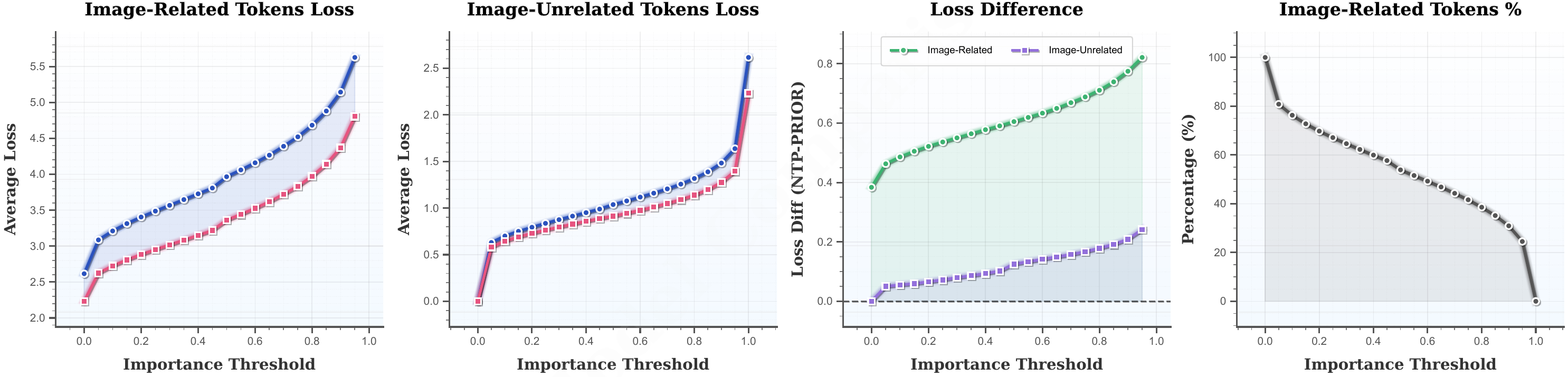}
  \vspace{-10pt}
\includegraphics[width=0.45\textwidth]{figs/raw_figs/elegant_legend.pdf}
\caption{\label{fig:loss_compare}\textbf{The comparison of NTP and \method regarding the achieved loss on image-related and image-unrelated tokens.} 
These two token groups are dynamically categorized based on varying the importance threshold $w_i$ (\eref{eq:eq3}).
We find that \method accelerates the LVLMs training, consistently achieving lower loss on both image-related and image-unrelated tokens.}
  \vspace{-18pt}
\end{figure}

\subsection{Comparative Loss Analysis of NTP and \method}
\label{sec:loss_analysis}
We compare NTP and \method from the loss perspective. 
By applying various importance thresholds, we categorize tokens into image-related and image-unrelated subsets based on their assigned importance scores $w_i$. 
We then measure the average NTP loss of H-LVLMs trained via two methods on these two distinct subsets.
The results presented in~\fref{fig:loss_compare} reveal a significant pattern: \method deliberately guides models to prioritize image-related tokens, achieving consistently lower loss on these information-rich elements. 
Interestingly, \method also optimizes learning on image-unrelated tokens, though this performance difference is less pronounced than for image-related tokens. 
This suggests that \method introduces general model improvement and effectively accelerates LVLMs training overall.
Furthermore, our analysis reveals a compelling trend: as we progressively increase the importance threshold, the performance gap between methods widens substantially. This monotonic relationship confirms that \method delivers increasingly significant improvements for tokens with higher image relevance, validating its fundamental design principle of prioritizing tokens that carry the most visually related information.

\subsection{Ablation Study}
\looseness=-1
We conduct an ablation study to understand the influence of $\alpha$ and $k$ in~\eqref{eq:eq2}.
While \method sets $\alpha$ to 1 by default, we experiment with $\alpha = 0.5, 2, 4$ to test sensitivity. 
Higher $\alpha$ increases focus on optimizing loss for image-related tokens identified by the reference model. 
Additionally, we evaluate the necessity of $k$, which maintains the token loss scale, by comparing performance with and without $k$ while fixing $\alpha = 1$. 
All experiments are conducted on H-LVLMs.
Results in~\fref{fig:ablation} (\sref{app:additional_exps}) demonstrate that performance degrades as $\alpha$ increases beyond 1 (to 2 or 4), yet remains stable within the moderate range of 0.5-1. Furthermore, removing $k$ negatively impacts performance on several benchmarks, validating our design choice to preserve the original loss scale. These findings suggest that a balanced approach to token weighting is crucial, as excessive emphasis on image-related tokens can impede the model's overall learning dynamics. Our empirical selection of $\alpha = 1$ and inclusion of scaling factor $k$ represents an optimal trade-off between focusing on visual content and maintaining stable training.

\section{Related Work}
\label{sec:related}

\looseness=-1
NTP loss on language tokens serves as the predominant pre-training objective for LVLMs~\citep{DBLP:conf/icml/WangYMLBLMZZY22, DBLP:journals/corr/abs-2305-06500, DBLP:journals/corr/abs-2308-12966, DBLP:journals/corr/abs-2403-05525, DBLP:journals/corr/abs-2409-11402, DBLP:journals/corr/abs-2407-06438}. 
Researchers have explored complementary approaches, including but not limited to:
(1) Contrastive or matching loss to align the vision-language modalities~\citep{DBLP:journals/corr/abs-1908-03557, DBLP:conf/nips/LuBPL19, DBLP:conf/icml/0001LXH22}.
(2) Distillation loss to expedite the training~\citep{DBLP:conf/nips/DiaoCLWLW24, liao2025multimodal}.
(3) Grounding objective to more effectively align the text tokens to the corresponding image regions~\citep{DBLP:conf/icml/KohSF23, DBLP:conf/cvpr/Rasheed0MS0CAX024}.
(4) Reconstructing the image based on the pre-defined codebook or external decoders~\citep{DBLP:conf/iclr/SunYCZZWGL0W24, DBLP:conf/cvpr/ZouDYGLLDBWYPWL23, DBLP:conf/iclr/GeZZGLWS24}.
(5) External constrains to improve the visual perception in LVLMs~\citep{DBLP:journals/corr/abs-2410-09575, DBLP:journals/corr/abs-2405-15232, chen2023vistruct}.
These supplementary objectives, though effective, compromise the simplicity of NTP in the original pre-training approach, complicating large-scale, efficient implementation~\citep{DBLP:journals/corr/abs-1909-08053, DBLP:conf/kdd/RasleyRRH20, DBLP:journals/corr/abs-2410-06511, DBLP:journals/corr/abs-2310-02239}. 
\method prioritizes maintaining NTP simplicity by merely adding a regularization term (\textit{i.e.,} a weighting factor) to each token during training.
In addition, \method avoids imposing human priors on the training dataset and remains compatible with advancements across all vision-language datasets (additional related work on LVLMs and vision-language training corpus are presented in~\sref{app:related}).


\section{Conclusion}
This work introduces \method, an advanced vision-language pre-training method that prioritizes the loss optimization on image-related tokens that a text-only reference model struggles to predict.
Our experiments demonstrate that \method significantly outperforms NTP, achieving 19\% and 8\% average relative improvement when implemented on LVLMs with and without pre-trained visual encoders, respectively.
Furthermore, \method exhibits more predictable and reliable scaling behaviors given increasing compute,
indicating that \method represents a promising pre-training algorithm.
We discuss the limitations and broader impact of \method in~\sref{app:limitation}.

%
%
%

\bibliography{main}
\bibliographystyle{plainnat}
\appendix
\newpage
\begin{figure}[t!]
  \centering
  \begin{minipage}[t]{0.33\textwidth}
    \centering
    \includegraphics[width=\textwidth]{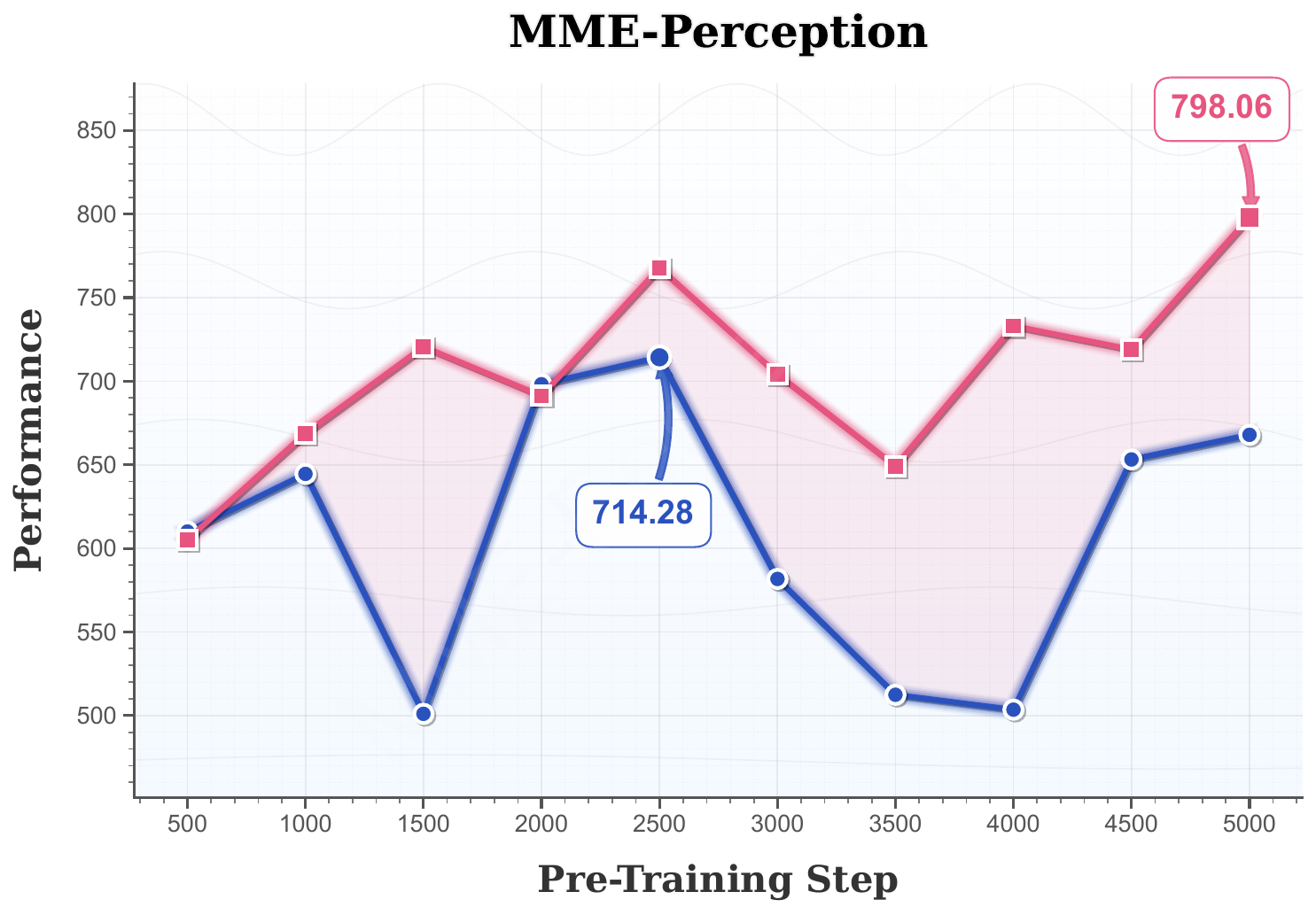}
    \label{fig:x1}
  \end{minipage}\hfill
  \begin{minipage}[t]{0.33\textwidth}
    \centering
    \includegraphics[width=\textwidth]{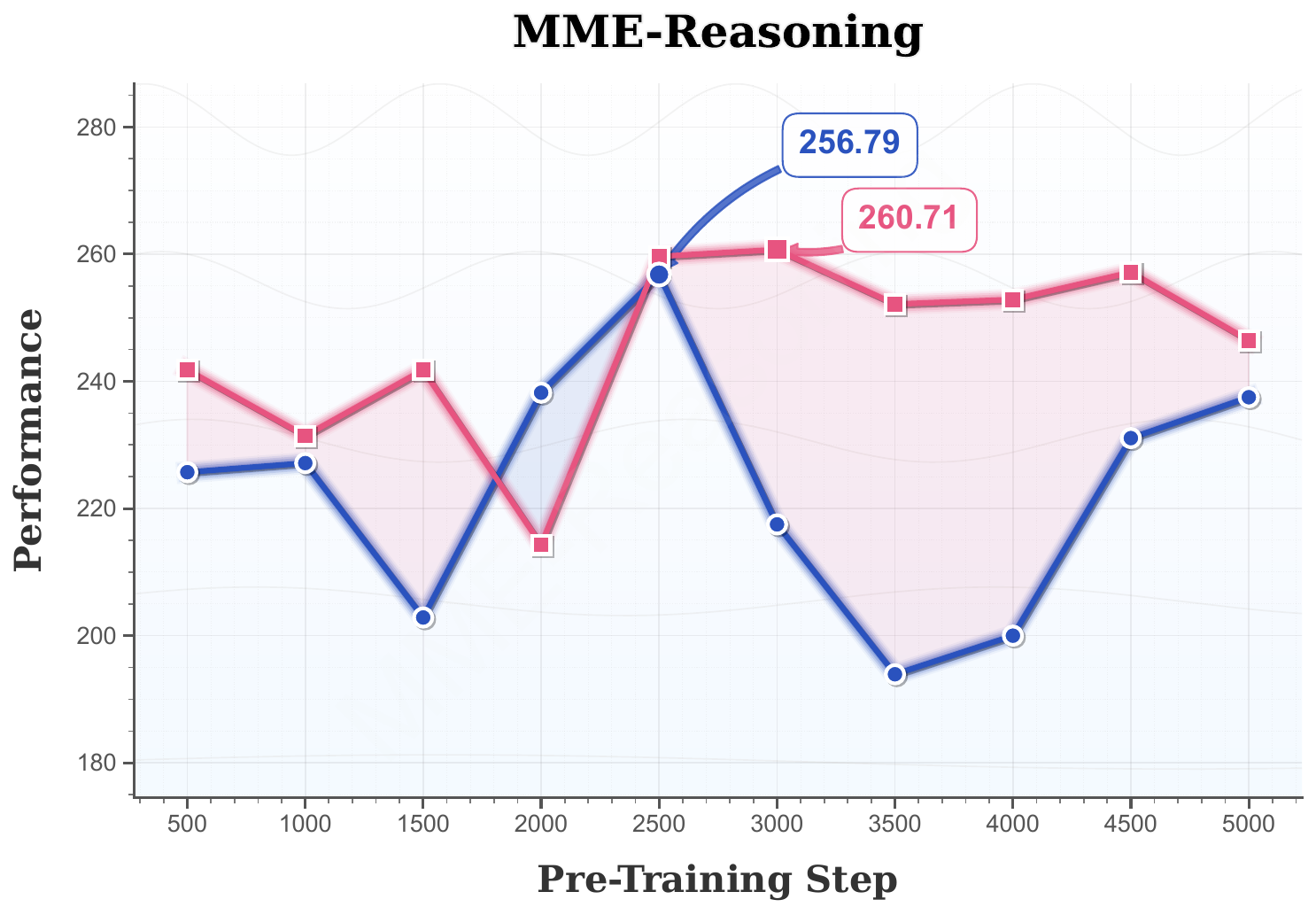}
    \label{fig:x2}
  \end{minipage}\hfill
  \begin{minipage}[t]{0.33\textwidth}
    \centering
    \includegraphics[width=\textwidth]{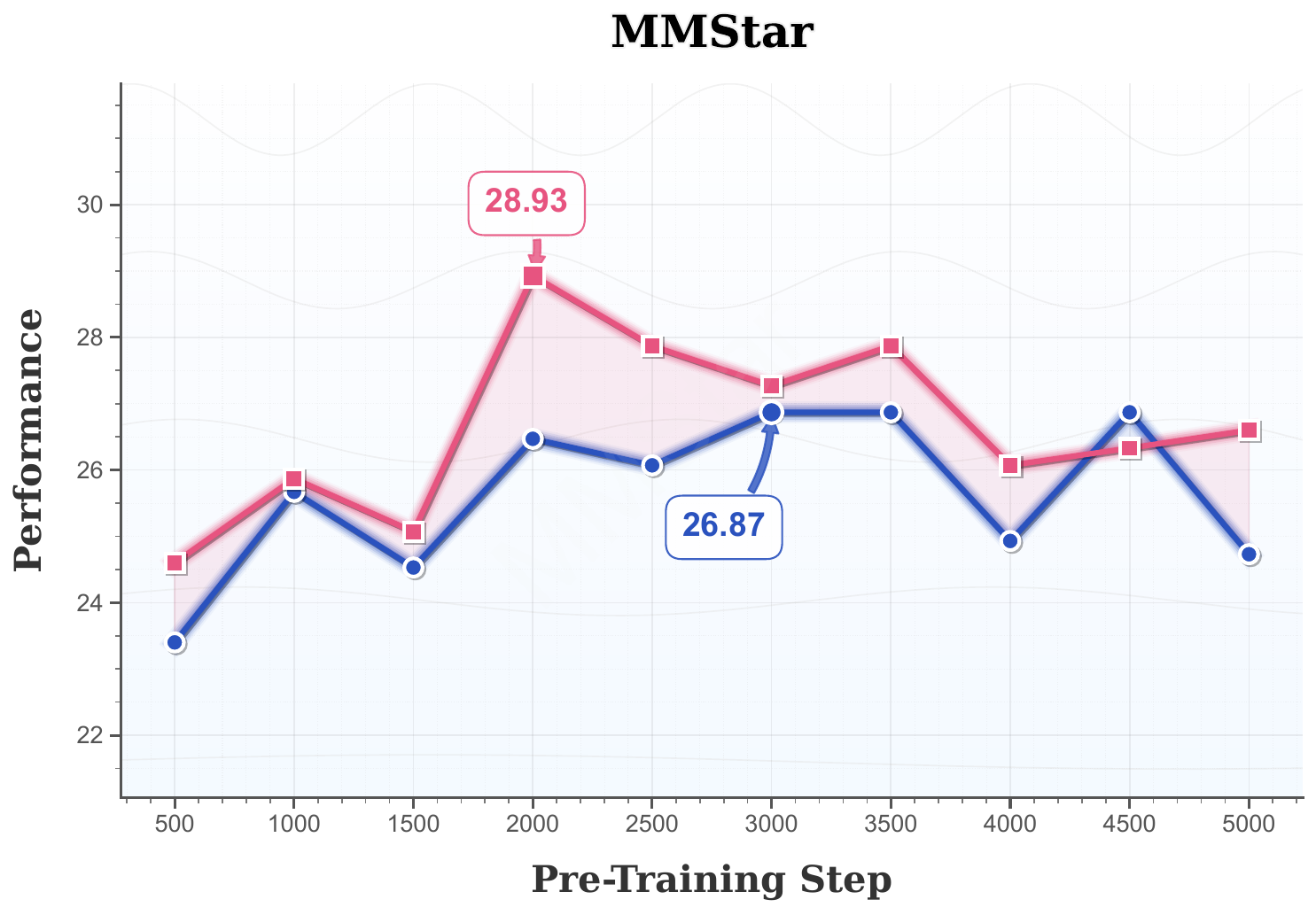}
    \label{fig:x3}
  \end{minipage}\hfill
  \vspace{-3pt}
  \begin{minipage}[t]{0.33\textwidth}
    \centering
    \includegraphics[width=\textwidth]{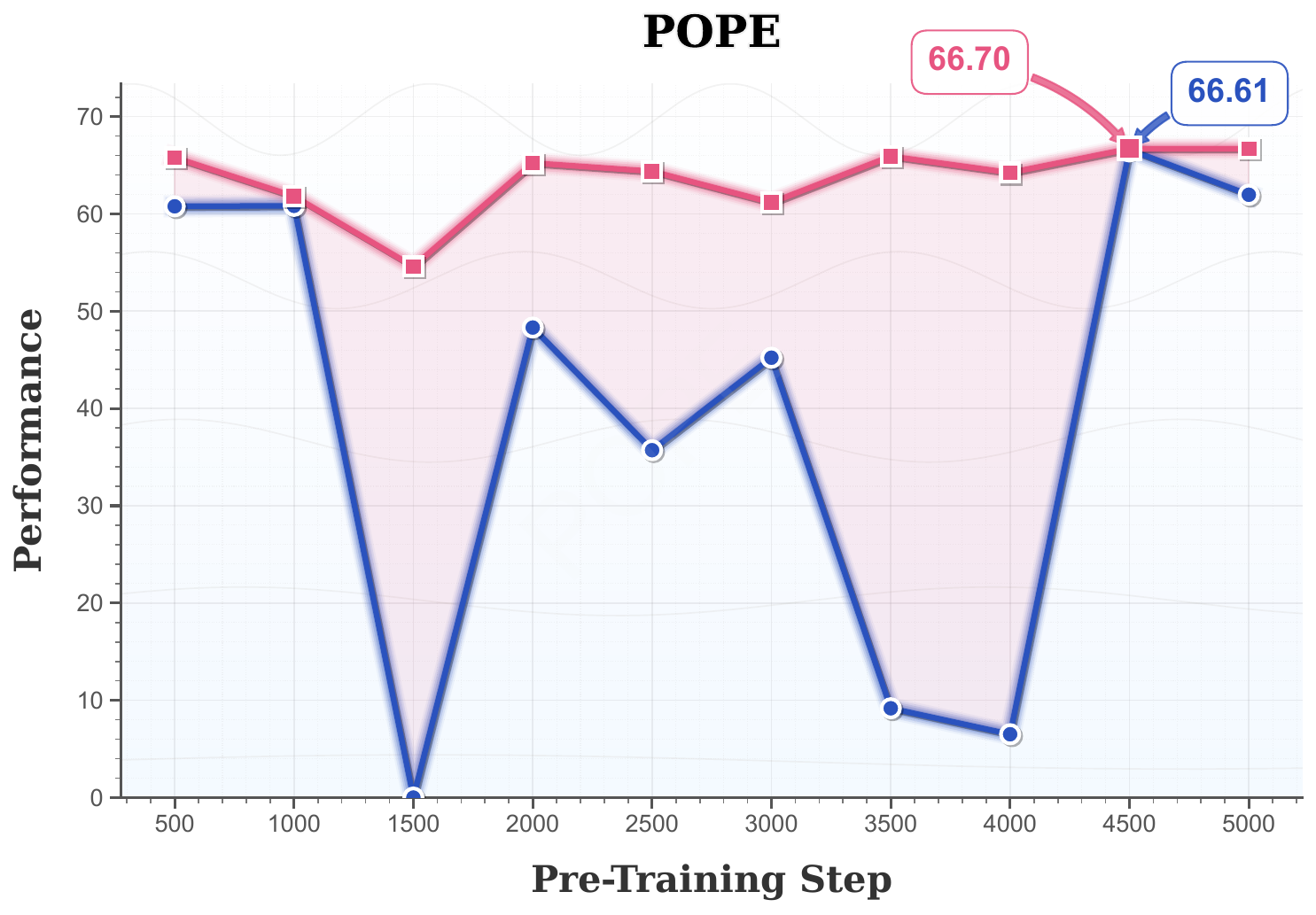}
    \label{fig:x1}
  \end{minipage}\hfill
  \begin{minipage}[t]{0.33\textwidth}
    \centering
    \includegraphics[width=\textwidth]{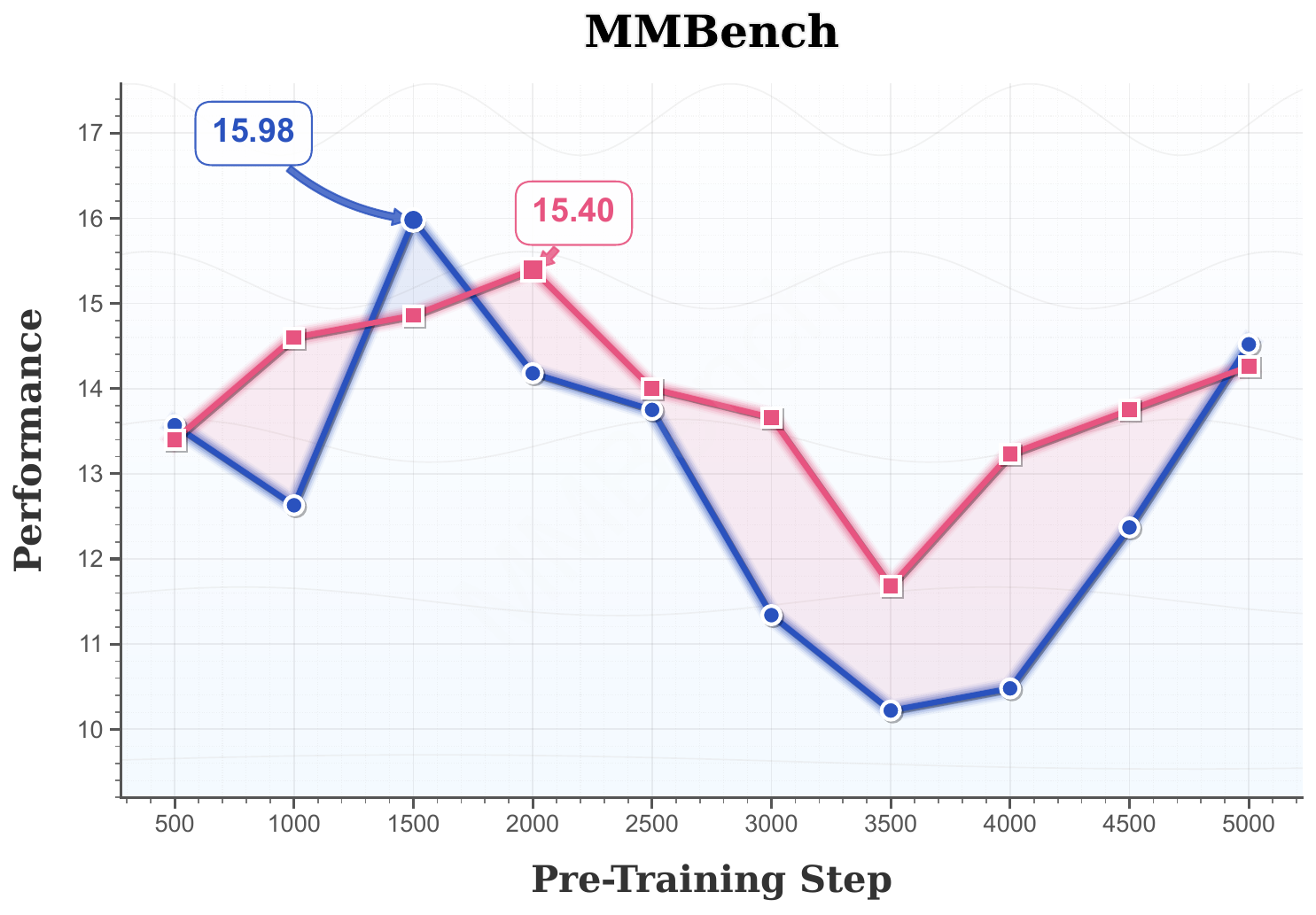}
  \end{minipage}\hfill
  \begin{minipage}[t]{0.33\textwidth}
    \centering
    \includegraphics[width=\textwidth]{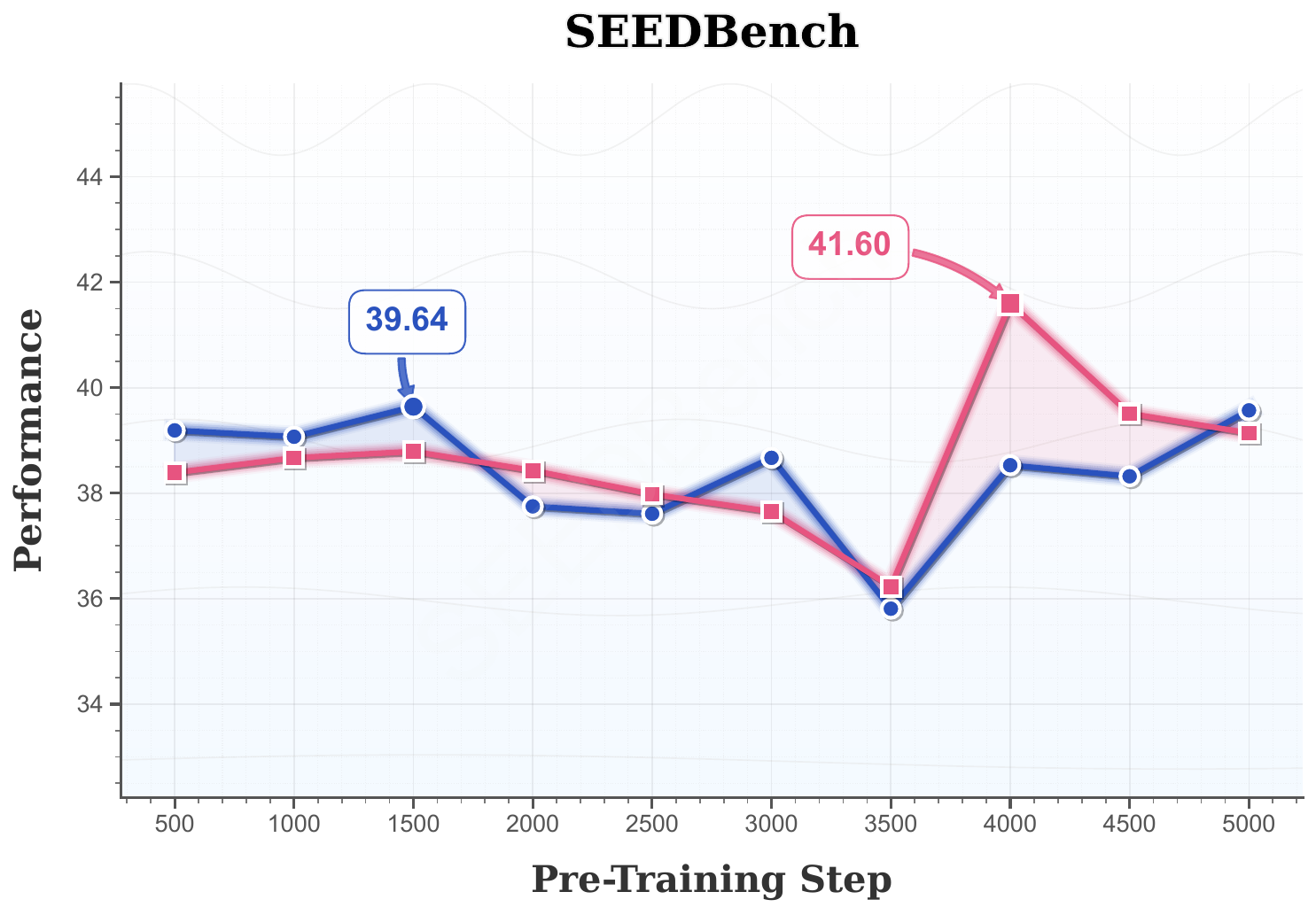}
  \end{minipage}\hfill
  \vspace{-10pt}
\includegraphics[width=0.45\textwidth]{figs/raw_figs/elegant_legend.pdf}
  \vspace{-8pt}
\caption{\label{fig:main_S}\textbf{Main experimental results of LVLMs with unified architectures.} We compare \method with the NTP vision-language pre-training on LVLMs with unified architectures.}
\end{figure}

 \vspace{-10pt}
 

\begin{figure*}[h!]
\centering
\includegraphics[width=0.85\textwidth]{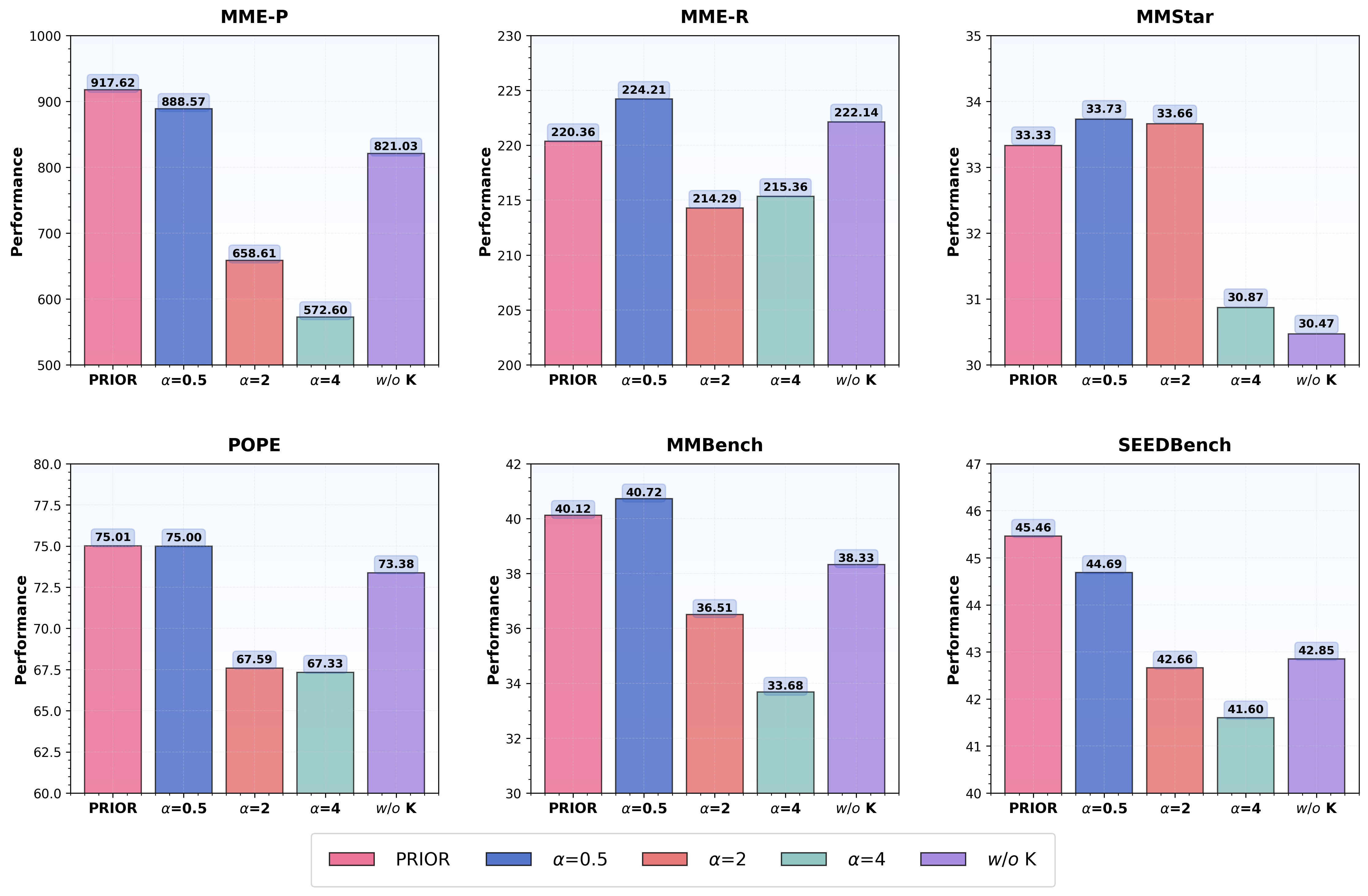}
 \caption{\label{fig:ablation}\textbf{The ablation study of \method regarding $\alpha$ and $k$ on H-LVLMs.} Results show optimal performance with $\alpha$=1 and scaling factor $k$, justifying the design choices in \method.}
 \end{figure*}
 \vspace{-10pt}

\section{Additional Experimental Results}
\label{app:additional_exps}
The experimental results of LVLMs with unified architectures on all benchmarks are shown in~\fref{fig:main_S}. 
The ablation study results are shown in~\fref{fig:ablation}.

\section{Additional Related Work}
\label{app:related}

\subsection{Large Vision-Language Models}
Existing research advances the development of LVLMs capable of addressing diverse tasks via a unified interface that can directly generate natural language, thus avoiding task-specific modifications~\citep{wang2021simvlm, DBLP:conf/icml/WangYMLBLMZZY22, DBLP:journals/corr/abs-2301-12597, DBLP:journals/corr/abs-2410-07073, luo2025openomni}.
Utilizing advanced pre-trained LLMs~\citep{gpt3, DBLP:journals/corr/abs-2303-12712, DBLP:journals/corr/abs-2407-21783, DBLP:journals/corr/abs-2412-15115} as the language component~\citep{DBLP:journals/corr/abs-2304-08485, DBLP:journals/corr/abs-2304-10592}, the instruction-following and complex reasoning abilities of LVLMs are significantly improved~\citep{DBLP:conf/ijcai/DuLLZ22, DBLP:journals/corr/abs-2404-07214}. 
Typically, LVLMs leverage extensive image-caption pair datasets~\citep{lin2014microsoft, schuhmann2021laion, schuhmann2022laion} to train a projector that maps image features into the embedding space of LLMs, thereby aligning the two modalities~\citep{DBLP:journals/corr/abs-2304-08485, DBLP:journals/corr/abs-2304-10592, flamingo, DBLP:journals/corr/abs-2301-12597, DBLP:journals/corr/abs-2303-05342, DBLP:conf/nips/AwadallaXLSLGSA24}.
Furthermore, large-scale vision-language instruction tuning datasets~\citep{su2023pandagpt, wei2023instructiongpt, liu2023aligning, gong2023multimodal, gao2023llama, li2023otter, DBLP:journals/corr/abs-2501-14818} and feedback datasets~\citep{chen2023dress, li2023silkie, sun2023aligning, DBLP:journals/corr/abs-2412-08687, zhao2025omnialign} are utilized to align LVLMs with human preferences, ensuring their ability to comprehend instructions and generate responses that are user-friendly.
In this work, we present a simple vision-language pre-training algorithm applicable to existing datasets that enhances visual-related training outcomes.

\subsection{Vision-Language Training Corpus}
The quality of training data is a critical determinant of LVLMs' ultimate performance.
The established efforts on improving the training data include but not limited to: 
(1) Distilling knowledge from advanced closed-source LVLMs~\citep{DBLP:conf/eccv/ChenLDZHWZL24, DBLP:conf/nips/0016WLD0ZCDB00024, DBLP:journals/corr/abs-2407-15838}, like GPT-4o~\citep{DBLP:journals/corr/abs-2410-21276}.
(2) Scaling the size of the dataset with scalable data collection pipelines~\citep{DBLP:journals/corr/abs-2406-08418, DBLP:conf/nips/AwadallaXLSLGSA24, DBLP:journals/corr/abs-2412-05243, wang2025scaling}.
(3) Incorporating the human-written filtering rules or pipelines~\citep{DBLP:journals/corr/abs-2412-05237, gohari2025gneissweb}.
(4) Curating domain-specific corpus to facilitate certain abilities in LVLMs~\citep{DBLP:conf/acl/0039WXWFK024, DBLP:conf/nips/YunLTBWJDWTL0NB24, DBLP:journals/corr/abs-2407-14506, DBLP:journals/corr/abs-2409-12568,  yang2025scaling}.
(5) Relying on the self-evolving ability in LVLMs~\citep{DBLP:journals/corr/abs-2412-17451, DBLP:journals/corr/abs-2409-05840}.
While existing work makes valuable progress toward higher-quality training data, these approaches face fundamental scalability constraints imposed by their data creation components, including the inherent capabilities of teacher models, pre-defined filtering criteria, and established pipelines—all of which create practical upper bounds on quality improvement.

\section{Theoretical Justification via Mutual Information}
\label{app:mutual}
The efficacy of \method can be further understood through the lens of mutual information theory. 
When optimizing vision-language models, we aim to maximize the predictive power of visual information $v$ with respect to text $t$. 
This can be formalized by maximizing the mutual information between $v$ and $t$, quantifying the uncertainty reduction about $t$ when $v$ is observed.
For data sampled from the joint distribution $(v,t)$, the mutual information is expressed as:
\begin{align}
I(v;t) &= \mathbb{E}_{(v,t)} \left[ \ln \frac{p(v,t)}{p(v)p(t)} \right] =\mathbb{E}_{(v,t)} \left[ \ln \frac{p(t|v)}{p(t)} \right]
\end{align}

Decomposing this expression at the token level, we obtain:
\begin{align}
I(v;t) = \mathbb{E}_{(v,t)} \left[ \sum_i \ln p(t_i|v,t_{<i}) - \ln p(t_i|t_{<i}) \right]
\end{align}

This formulation shows that mutual information is maximized when there's a large discrepancy between token probability given both visual and textual context $p(t_i|v,t_{<i})$ versus textual context alone $p(t_i|t_{<i})$. Tokens exhibiting this difference benefit most from visual inputs.


The weighting mechanism in \method, defined as $w_i = (1-p_r(t_i|t_{<i}))^\alpha$, implicitly aligns with this mutual information objective. 
By assigning higher importance scores to tokens that are difficult to predict from text alone,
\method effectively prioritizes tokens where visual information potentially provides the greatest reduction in uncertainty. This approach creates a natural emphasis on tokens where $\ln p(t_i|v,t_{<i}) - \ln p(t_i|t_{<i})$ is likely to be large, thus indirectly promoting higher mutual information between vision and language representations.

This intuitively corresponds to focusing loss optimization on the tokens where the LVLMs have the greatest opportunity to outperform text-only predictions by leveraging visual information. 
In essence, \method adaptively modulates the learning signal based on the potential information gain from incorporating visual context, leading to more efficient and effective vision-language pre-training.






\section{Limitations and Broader Impacts}
\label{app:limitation}
\paragraph{Limitations}
\method requires the LVLMs and the reference LLM to share the same tokenizer, preventing the creation of a universal dataset for all LVLMs pre-training. 
This constraint on the tokenizer necessitates specific implementations and processing, increasing computational overhead when deploying across multiple model architectures.

\paragraph{Broader impacts}
\method's token prioritization approach significantly enhances LVLMs' performance across benchmarks, potentially accelerating progress in building advanced general-purpose LVLMs. 
Additionally, by reducing hallucination risk through better grounding of language in visual content, \method contributes to developing more trustworthy LVLMs models.
However, there remains the possibility of misuse, and the differential weighting approach may unintentionally amplify existing biases in training data. 
We encourage continued research into responsible deployment practices and bias mitigation techniques alongside performance improvements.

\end{document}